\documentclass[11pt,letterpaper]{article}

\usepackage[T1]{fontenc}
\usepackage[utf8]{inputenc}
\usepackage{mathptmx}          
\usepackage{helvet}           
\usepackage{courier}
\usepackage{textcomp}
\usepackage{amsmath,amssymb}
\usepackage{microtype}

\usepackage[letterpaper,top=0.92in,bottom=0.95in,left=0.95in,right=0.95in]{geometry}

\usepackage{xcolor}
\definecolor{arxivblue}{RGB}{46,90,168}     
\definecolor{arxivink}{RGB}{24,32,48}      
\definecolor{arxivrule}{RGB}{176,190,210}  
\definecolor{arxivlink}{RGB}{46,90,168}    
\definecolor{arxivgray}{RGB}{96,108,126}   
\definecolor{absbg}{RGB}{243,246,251}      

\usepackage{graphicx}
\usepackage{booktabs,longtable,array}
\usepackage[font=small,labelfont={bf,color=arxivblue},labelsep=period,skip=6pt]{caption}
\usepackage[font=small,skip=4pt]{subcaption}
\usepackage{placeins}

\usepackage[shortlabels,inline]{enumitem}

\usepackage[numbers,sort&compress]{natbib}

\usepackage[colorlinks,breaklinks,allcolors=arxivlink,citecolor=arxivblue,
            urlcolor=arxivlink,linkcolor=arxivblue]{hyperref}
\hypersetup{
  pdftitle={When Figures Disagree with Text: Evaluating Autonomous Visual Evidence Verification in Scientific Papers},
  pdfauthor={Guijia Zhang, Hongbo Qiu, Jiajun Zha, Guohao Cai, Hao Zheng, Huihao Jing, Kexu Cheng, Shipeng Luo, JinYang Huang, Shing-Chi Cheung, Changsheng Lu, Haoxuan Che, Harry Yang},
  pdfsubject={Multimodal document understanding; figure--text consistency verification},
  pdfkeywords={multimodal LLMs, document understanding, figure--text consistency, autonomous verification, PRISMM-Bench}
}

\input{glyphtounicode}
\providecommand{\PENDING}[1]{{\color{red}\ensuremath{\blacktriangle}}}
\newcommand{\RAnoprompt}{4.30}

\newcommand{\RAtarget}{20.43}

\newcommand{\RAnopromptCI}{{[1.08, 8.06]}}

\newcommand{\RAtargetCI}{{[13.44, 28.49]}}

\newcommand{\RAtgtNoFrame}{$+16.13$}
\newcommand{\RAtgtNoFrameCI}{{[$+9.14$, $+24.19$]}}

\newcommand{\RAframeNoTarget}{$+2.69$}
\newcommand{\RAframeNoTargetCI}{{[$-3.23$, $+9.14$]}}
\newcommand{\RAframeTarget}{$-3.23$}
\newcommand{\RAframeTargetCI}{{[$-10.22$, $+3.23$]}}
\newcommand{\RAinteract}{$-5.91$}
\newcommand{\RAinteractCI}{{[$-15.59$, $+2.15$]}}

\newcommand{\RArecallThree}{34.41}
\newcommand{\RArecallThreeCI}{{[23.66, 46.24]}}
\newcommand{\RArecallFive}{57.53}
\newcommand{\RArecallFiveCI}{{[43.55, 71.51]}}
\newcommand{\RAtwoStage}{4.84}
\newcommand{\RAtwoStageCI}{{[1.61, 8.60]}}
\newcommand{\RArecovered}{3.3}
\newcommand{\RArecoveredCI}{{[$-33.30$, $+27.30$]}}
\newcommand{\RAtokenRatio}{2.04}
\newcommand{\RAtwoStageMinusBase}{$+0.54$}
\newcommand{\RAtwoStageMinusBaseCI}{{[$-3.76$, $+5.38$]}}

\newcommand{\RAtwoStageMinusTarget}{$-15.59$}
\newcommand{\RAtwoStageMinusTargetCI}{{[$-22.58$, $-9.68$]}}
\newcommand{\RAsTwoHitThree}{6.25}
\newcommand{\RAsTwoHitThreeCI}{{[0.00, 13.85]}}
\newcommand{\FUNrows}{384}
\newcommand{\FUNpapers}{353}
\newcommand{\FUNpdfRows}{204}
\newcommand{\FUNpdfPapers}{191}
\newcommand{\FUNvisRows}{198}
\newcommand{\FUNvisPapers}{186}
\newcommand{\FUNcand}{186}
\newcommand{\FUNenrolled}{31}
\newcommand{\FUNnotenrolled}{155}
\newcommand{\FUNmedPages}{17}
\newcommand{\FUNmedPagesOther}{15}
\newcommand{\FUNmeanPages}{19.0}
\newcommand{\FUNmeanPagesOther}{16.1}
\newcommand{\FUNpermP}{0.009}
\newcommand{\FUNfigOnly}{29}
\newcommand{\FUNfigOnlyOther}{112}
\newcommand{\FUNtableOnly}{0}
\newcommand{\FUNtableOnlyOther}{70}

\newcommand{\MVcohortGen}{2026-11-03}
\newcommand{\MVcohortGrade}{2026-11-03 to 2026-11-04}
\newcommand{\MVprobeGen}{2026-11-12 to 2026-11-15}
\newcommand{\MVprobeGrade}{2026-11-12 to 2026-11-15}
\newcommand{\MVadjud}{2026-11-04 to 2026-11-12}
\newcommand{\MVcohortDated}{1,659}
\newcommand{\MVcohortAlias}{1}
\newcommand{\MVprobeDated}{1,194}
\newcommand{\MVprobeAlias}{1,104}
\newcommand{\MVgraderAliasMin}{4.3}
\newcommand{\MVgraderAliasMax}{100.0}
\newcommand{\MVnDated}{1,020}
\newcommand{\MVagreeDated}{93.6}
\newcommand{\MVnAlias}{923}
\newcommand{\MVagreeAlias}{94.8}
\newcommand{\MVbaseDatedP}{1}
\newcommand{\MVbaseDatedN}{53}
\newcommand{\MVbaseDatedPct}{1.9}
\newcommand{\MVbaseAliasP}{7}
\newcommand{\MVbaseAliasN}{133}
\newcommand{\MVbaseAliasPct}{5.3}
\newcommand{\MVtgtDatedP}{20}
\newcommand{\MVtgtDatedN}{121}
\newcommand{\MVtgtDatedPct}{16.5}
\newcommand{\MVtgtAliasP}{18}
\newcommand{\MVtgtAliasN}{65}
\newcommand{\MVtgtAliasPct}{27.7}
\newcommand{\HAkappaFlaw}{0.898}
\newcommand{\HAkappaBinary}{0.898}

\newcommand{\HAagree}{90.8}
\newcommand{\HAkappaHuman}{0.642}
\newcommand{\HAprec}{52.6}
\newcommand{\HArec}{100.0}
\newcommand{\HAn}{100}
\newcommand{\HAoffkey}{1.4}
\newcommand{\HAtargetMiss}{9/30 (30.0\%)}
\newcommand{\HAoffkeyCI}{{[0.0, 4.5]}}

\renewenvironment{figure*}[1][]{\begin{figure}[#1]}{\end{figure}}
\renewenvironment{table*}[1][]{\begin{table}[#1]}{\end{table}}

\makeatletter
\renewcommand{\section}{\@startsection{section}{1}{\z@}%
  {-2.4ex plus -1ex minus -.2ex}{1.2ex plus .2ex}%
  {\normalfont\large\bfseries\color{arxivblue}}}
\renewcommand{\subsection}{\@startsection{subsection}{2}{\z@}%
  {-1.9ex plus -1ex minus -.2ex}{0.9ex plus .2ex}%
  {\normalfont\normalsize\bfseries\color{arxivblue}}}
\makeatother

\newcommand{\affilmark}[1]{\textsuperscript{#1}}

\newcommand{\corrmark}{\textsuperscript{\textcolor{arxivblue}{$\boxtimes$}}}
\newcommand{\coremark}{\textsuperscript{\textcolor{arxivblue}{$\bigstar$}}}
\newcommand{\leadmark}{\textsuperscript{\textcolor{arxivblue}{$\blacklozenge$}}}
\newcommand{\dispfont}{\sffamily\bfseries}
\newcommand{\dispplain}{\sffamily}

\newsavebox{\abstractpanel}
\renewenvironment{abstract}{%
  \par\vspace{7pt}%
  \begin{lrbox}{\abstractpanel}%
  \begin{minipage}{\dimexpr\textwidth-2\fboxsep\relax}%
    \vspace{9pt}%
    \noindent{\dispfont\color{arxivblue}\fontsize{11}{13}\selectfont Abstract}\par
    \vspace{5pt}%
    \small\noindent\ignorespaces
  }{%
    \par\vspace{9pt}%
  \end{minipage}%
  \end{lrbox}%
  \noindent\colorbox{absbg}{\usebox{\abstractpanel}}\par
  \vspace{6pt}%
}

\begin{document}
\thispagestyle{plain}
\pagestyle{plain}

\vspace*{-2pt}


\noindent\begin{center}%
  \vspace{3pt}%
  \includegraphics[height=2.7em]{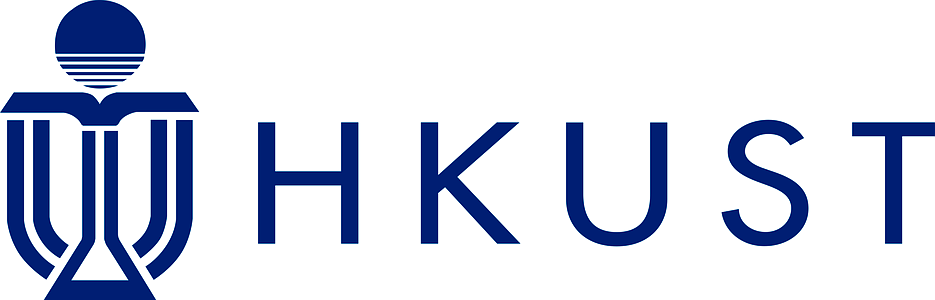}%
  \hspace{0.6em}%
  \includegraphics[height=2.7em]{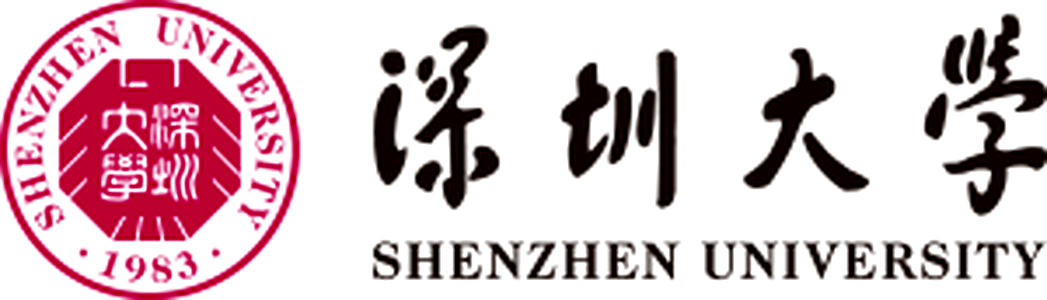}%
  \hspace{0.6em}%
  \includegraphics[height=2.9em]{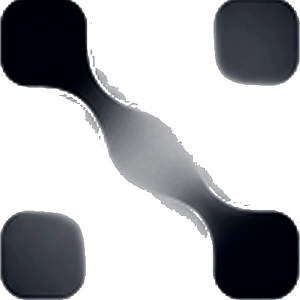}%
  \hspace{0.6em}%
  \includegraphics[height=2.9em]{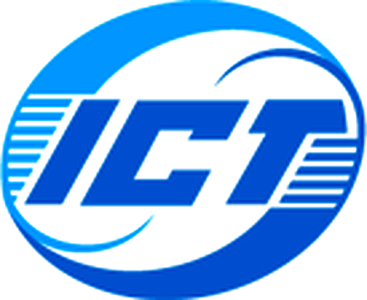}%
  \hspace{0.6em}%
  \includegraphics[height=2.9em]{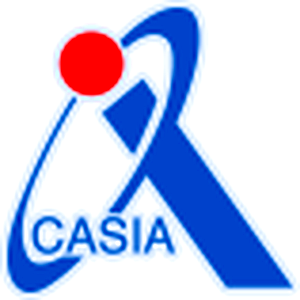}%
  \hspace{0.6em}%
  \includegraphics[height=2.7em]{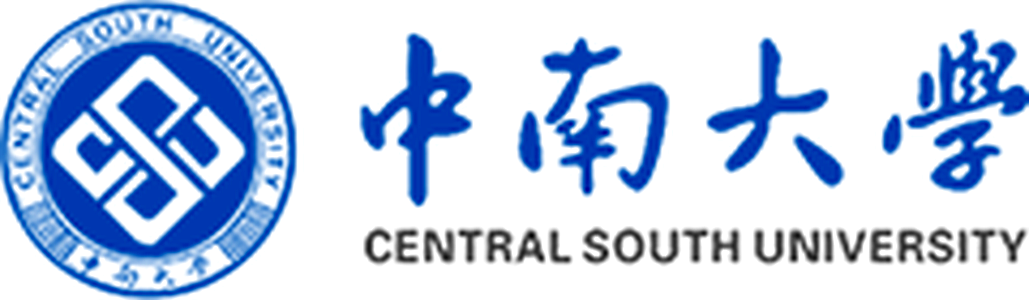}%
  \vspace{2pt}%
\end{center}\par

\vspace{2pt}
\noindent{\color{arxivblue}\rule{\textwidth}{2pt}}\par
\vspace{9pt}

{\dispfont\color{arxivink}\fontsize{21}{25}\selectfont
\noindent When Figures Disagree with Text\par}
\vspace{4pt}
{\dispplain\color{arxivink}\fontsize{14}{18}\selectfont
\noindent Evaluating Autonomous Visual Evidence Verification\par
\noindent in Scientific Papers\par}

\vspace{11pt}

{\dispfont\color{arxivink}\fontsize{11}{15}\selectfont
\noindent
\begin{tabular}{@{}l@{\quad}l@{\quad}l@{\quad}l@{\quad}l@{}}
  Guijia Zhang\coremark\affilmark{1} & Hongbo Qiu\coremark\affilmark{2} & Jiajun Zha\affilmark{1} &
  Guohao Cai\leadmark\affilmark{3} & Hao Zheng\affilmark{2} \\
  Huihao Jing\affilmark{1} & Kexu Cheng\affilmark{4} & Shipeng Luo\affilmark{5} &
  JinYang Huang\affilmark{6} & \\
  Shing-Chi Cheung\affilmark{1} & Changsheng Lu\leadmark\affilmark{1} &
  Haoxuan Che\affilmark{3}\corrmark & Harry Yang\affilmark{1}\corrmark &
\end{tabular}\par}

\vspace{2pt}
{\dispplain\fontsize{8.5}{11}\selectfont
\noindent{\color{arxivblue}\coremark}\enspace Core contributor.\qquad
{\color{arxivblue}\leadmark}\enspace Project leads.\qquad
{\color{arxivblue}\corrmark}\enspace Corresponding authors.\par}

\vspace{7pt}

{\dispplain\fontsize{10.5}{14}\selectfont
\noindent
\begin{tabular}{@{}r@{\,}l@{}}
  \affilmark{1} & The Hong Kong University of Science and Technology \\
  \affilmark{2} & Shenzhen University \\
  \affilmark{3} & XGEN Labs \\
  \affilmark{4} & Institute of Computing Technology, Chinese Academy of Sciences \\
  \affilmark{5} & Institute of Automation, Chinese Academy of Sciences \\
  \affilmark{6} & Central South University
\end{tabular}\par}

\vspace{4pt}

\begin{abstract}
Scientific figures carry evidence for the claims that cite them, so reviewing a paper means checking that a figure agrees with the text, and not only describing what the figure shows. Benchmarks of figure--text verification typically hand the model the figure, the claim, or the knowledge that an inconsistency exists, which leaves untested whether a model can find the figure worth checking in a complete paper, relate a visual element to the claim it conflicts with, and recognize the conflict. We study this autonomous setting on 31 full manuscripts, each carrying a reviewer-documented figure--text inconsistency from PRISMM-Bench that is hidden from the model. Six multimodal models review each PDF under a neutral prompt, and we then add one input at a time to the same manuscripts. Of 1{,}445 open-ended reviews, 60.6\% mention the flawed figure, yet a grading rubric accepts only 3.3\% as diagnosing its contradiction; a blind audit finds the rubric lenient, so we read these as rubric-accepted rates and compare conditions rather than absolute values. In a separate controlled run, naming the figure, without stating that it is flawed, raises rubric-accepted diagnosis from 4.3\% to 20.4\%, while a consistency instruction and coarse visual cues do not. When the contradiction is visible only in the rendered figure, text-only review yields no accepted diagnosis, while review of the rendered pages with the figure named reaches 15.8\%. A baseline that lets the model choose which figures to check, ranked by centrality, stays at 4.8\%. Deciding what to verify is thus a large part of the difficulty, and the selection rule we tested does not resolve it. Multimodal document understanding should be evaluated on whether a model can propose what to verify, not only answer questions about a figure it is handed.
\par\vspace{5pt}
\noindent{\color{arxivblue}\textbf{Keywords:}}\quad multimodal large language models; document understanding; figure--text consistency; autonomous verification; peer review; PRISMM-Bench
\end{abstract}

\vspace{8pt}

\noindent
\begin{tabular}{@{}p{0.31\textwidth}@{\hspace{0.035\textwidth}}p{0.31\textwidth}@{\hspace{0.035\textwidth}}p{0.31\textwidth}@{}}
  \colorbox{absbg}{\parbox[c][52pt][c]{\dimexpr0.31\textwidth-2\fboxsep\relax}{%
    \centering\vspace{2pt}%
    {\dispfont\color{arxivblue}\fontsize{20}{22}\selectfont 60.6\%}\par
    \vspace{1pt}%
    {\dispplain\fontsize{7.5}{9.5}\selectfont mention the flawed figure\par}%
    \vspace{2pt}}} &
  \colorbox{absbg}{\parbox[c][52pt][c]{\dimexpr0.31\textwidth-2\fboxsep\relax}{%
    \centering\vspace{2pt}%
    {\dispfont\color{arxivblue}\fontsize{20}{22}\selectfont 3.3\%}\par
    \vspace{1pt}%
    {\dispplain\fontsize{7.5}{9.5}\selectfont are accepted as diagnosing it\par}%
    \vspace{2pt}}} &
  \colorbox{absbg}{\parbox[c][52pt][c]{\dimexpr0.31\textwidth-2\fboxsep\relax}{%
    \centering\vspace{2pt}%
    {\dispfont\color{arxivblue}\fontsize{20}{22}\selectfont 20.4\%}\par
    \vspace{1pt}%
    {\dispplain\fontsize{7.5}{9.5}\selectfont when the figure is named\par}%
    \vspace{2pt}}}
\end{tabular}\par

\vspace{8pt}

\section{Introduction}
\label{sec:intro}

\begin{figure*}[t]
\centering\includegraphics[width=\textwidth]{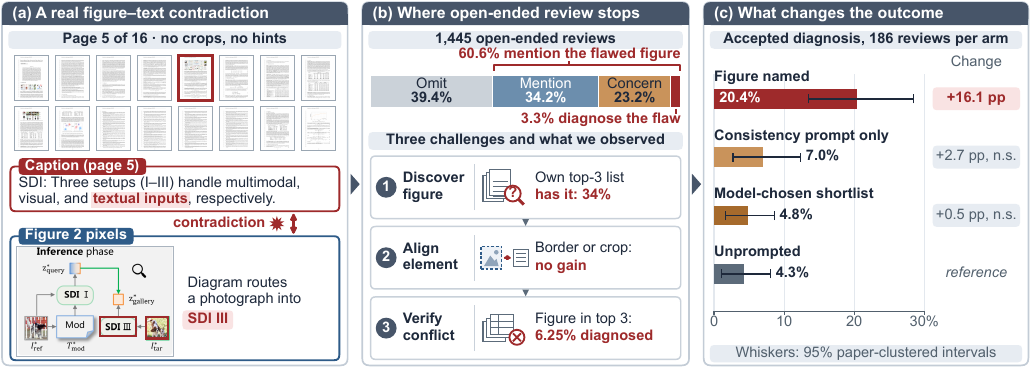}
\caption{\textbf{A figure--text contradiction that exists only between pixels and a caption clause, and where autonomous review stops.}
(a)~Page~5 of 16 in an unmodified manuscript. The caption assigns textual input to the third setup, but the diagram routes a photograph into it.
(b)~The bar shows the 1{,}445 open-ended reviews, which pool four input conditions including figure borders. The three cards give one observation for each challenge: shortlist recall (Section~\ref{sec:twostage}), the border and crop contrasts (Sections~\ref{sec:null} and~\ref{sec:factorial}), and the exploratory shortlist-subset rate (Section~\ref{sec:twostage}). They illustrate the challenges and do not measure them separately.
(c)~Rubric-accepted diagnosis in four arms of the controlled run (186 reviews each). Naming the figure helps, a consistency instruction alone does not, and a model-chosen shortlist stays near the unprompted level. Whiskers show 95\% paper-clustered intervals.}\label{fig:teaser}\end{figure*}

Scientific figures are evidence. A claim in the text is only as credible as the plot, table, or diagram cited in its support, so a careful reviewer does more than understand a figure: the reviewer checks that it shows what the text says. Multimodal language models are now fluent at the first task. They summarize plots, tables, and architecture diagrams, and they are already used to give feedback on real manuscripts~\cite{liang2023reviewgen,thakkar2026peerreview}. Whether they can do the second matters, because a figure that contradicts its text changes how a result should be read, while a fluent description of that figure gives no sign that it was compared with any claim. Figure~\ref{fig:teaser}a shows an authentic case from a 16-page manuscript. The caption of a pipeline diagram states that three setups handle multimodal, visual, and textual inputs, but the rendered diagram routes a photograph into the setup assigned to textual input. No sentence in the paper states this conflict. It exists only between a few pixels of the figure and one clause of the caption.

Finding such a conflict in a complete paper, with no hint that one exists, is harder than answering a question about a given figure, because it requires three steps. The reviewer must first decide which of many figures, tables, and claims deserves scrutiny (\textbf{candidate discovery}). Having chosen a figure, the reviewer must tie a specific visual element, such as an arrow, a legend entry, or a curve, to the clause elsewhere in the paper that makes a claim about it (\textbf{fine-grained evidence--claim alignment}). Finally, the reviewer must hold the element and the claim side by side and judge whether they agree, instead of writing coherent commentary on each (\textbf{contradiction verification}). These steps separate three settings that are often conflated. \emph{Figure understanding} asks what a figure shows. \emph{Guided verification} supplies the figure or the claim, or states that an inconsistency exists. Claim-verification benchmarks pair a claim with the figure or paper context it concerns~\cite{wang2025sciver,lal2025musciclaims,akhtar2024chartcheck}, and PRISMM-Bench, built from real reviewer comments, states that an inconsistency exists and asks the model to choose among answers~\cite{prismm}. \emph{Autonomous verification} supplies only the complete paper and asks for a review. Success in guided settings shows that a model can verify a stated claim or recognize a known inconsistency once the question is posed. It does not show that the model would pose the question itself, which is what a reviewer has to do. Our experiments change inputs and observe outputs, so they do not measure the three steps as separate failure mechanisms inside a model.

We evaluate autonomous verification through full-document review. The real inconsistencies that peer reviewers documented in PRISMM-Bench become hidden targets in 31 complete manuscripts. Six frontier multimodal models review each PDF under a neutral prompt, with no answer choices and no statement that an error exists, and we record how far each review gets: it may omit the flawed element, mention it, raise a concern about it, or be accepted as diagnosing the documented contradiction. Three studies on the same manuscripts follow. \textbf{RQ1:} where do unguided reviews stop? \textbf{RQ2:} which single change to the input carries them further, such as naming the figure, asking for a consistency check, highlighting the figure, or removing the visual channel? \textbf{RQ3:} can the model supply the most helpful input itself, by selecting candidate figures before it verifies them? The third study is a diagnostic baseline and not a proposed method. Together, the questions measure how far autonomous verification lies from guided verification.

The three studies build one argument, although they come from different runs. In the open-ended cohort of 1{,}445 reviews, 60.6\% mention the flawed figure while 3.3\% are accepted as diagnosing its contradiction, and the flawed figure is not clearly mentioned more often than a matched unflawed neighbor (RQ1). Reviews discuss figures far more than they check them against the text. In a separate controlled run with 186 reviews per arm, whose unprompted arm is accepted on 4.3\%, naming the figure, without stating that it is flawed, raises rubric-accepted diagnosis to 20.4\%, while a consistency instruction, whole-figure highlights, and moving the evidence beside the caption do not (RQ2). The help is real but partial: about four of five reviews still miss the contradiction, and many near-misses reach the right figure while asserting a different relation. When the model selects candidates itself in the same design, ranking figures by centrality, accepted diagnosis is 4.8\%, near the unprompted level, and its top-three shortlist contains the flawed figure in 34\% of runs (RQ3). Deciding what to check is thus a large part of the difficulty, and the one selection rule we tested does not resolve it. Every rate is a rubric-accepted rate; a blind audit found the rubric lenient, so we compare conditions rather than read absolute values as detection accuracy, and all of these are observations about review text, not about a model's internal process.

We make three contributions.
\begin{itemize}\setlength\itemsep{0pt}
\item \textbf{An evaluation of autonomous visual evidence verification.} A protocol that asks multimodal models to review complete PDFs containing hidden, reviewer-documented figure--text inconsistencies, scores each review on four nested outcomes, and checks the rubric by replay, an independent grader panel, and a blind human audit.
\item \textbf{Experimental evidence that separates the settings.} Discussing a figure is not verifying it, naming the figure is the one input change that raises rubric-accepted diagnosis substantially, and most contradictions stay undiagnosed even then. A centrality-based shortlist does not recover the benefit.
\item \textbf{Implications for multimodal document understanding.} Benchmarks that supply the figure or the existence of an error can overstate what a reviewer achieves without such help, so evaluations should also ask models to propose what to verify and to ground that choice in specific visual evidence.
\end{itemize}

\section{Related Work}
\label{sec:related}

Prior work on scientific verification differs mainly in what it gives the model: the claim, the figure, the knowledge that an error exists, or none of these. We organize the literature by that choice, because it determines which of the three steps in Section~\ref{sec:intro} a benchmark can measure.

\textbf{The claim and its evidence are given.} Scientific claim verification began with text, where a model retrieves abstracts that support or refute a claim and identifies the rationale sentences~\cite{wadden2019factcheck}. Multimodal versions move the evidence into charts and figures. ChartCheck verifies claims against real-world chart images~\cite{akhtar2024chartcheck}. MuSciClaims pairs each claim with a figure and its caption, and its diagnostic tasks show that models localize supporting evidence in a figure poorly and tend to judge claims as supported~\cite{lal2025musciclaims}. SciVer verifies claims over text, tables, and charts from one paper with expert-annotated evidence~\cite{wang2025sciver}, and SPIQA poses questions about figures in scientific papers~\cite{pramanick2024spiqa}. These benchmarks show that checking a stated claim against a figure is hard. Because the claim and the figure are supplied, they cannot show whether a model would select that claim and figure on its own.

\textbf{The existence of an inconsistency is given.} MMIR injects synthetic inconsistencies into webpages, slides, and posters, and finds that models handle conflicts between two elements better than conflicts confined to one element of a complex layout~\cite{yan2025mmir}. PRISMM-Bench is the closest work to ours. It collects 384 inconsistencies that peer reviewers flagged across text, figures, tables, and equations, and tests identification, remedy, and pair matching as multiple-choice tasks, with context ranging from a focused crop to the whole paper~\cite{prismm}. Models score low even though the question states that an inconsistency exists. We use its documented inconsistencies for a different measurement. The model receives the full manuscript and a neutral review request with no answer choices, and we score the review it writes. PRISMM-Bench asks whether a model recognizes the right answer among options. We ask whether it produces the diagnosis unaided, and our controlled arms restore one hint at a time, such as the identity of the figure or the conflicting clause.

\textbf{The target is given in the query.} Chart, table, and document benchmarks ask questions about a given figure or page~\cite{mathew2021docvqa,masry2022chartqa,wang2024charxiv}, and long-document and scientific suites extend these to many pages~\cite{ma2024mllongbench,wu2024mmsci}. Grounding benchmarks test whether a model can locate a small target on a dense screen~\cite{cheng2024seeclick,li2025screenspotpro}, and multi-image benchmarks test whether it can find a cue among many images~\cite{wu2024visualhaystacks}. In each case the query or the input identifies the target. Studies of model vision also report systematic failures on basic visual patterns~\cite{tong2024eyes}, and hallucination benchmarks study the opposite error, in which a model reports content that is absent~\cite{li2023hallu}. Verifying a figure against a claim needs both finding the relevant figure in a long document and grounding a specific element within it, without a query. In our setting the content is present and the contradiction goes unreported.

\textbf{Nothing is given except the paper.} Automated reviewing studies are the closest in setup, since the model receives a paper and writes a review. Work in this area has collected peer-review corpora~\cite{li2023peerread}, tested whether language models give useful feedback~\cite{liu2023reviewergpt,liang2023reviewgen}, run a large randomized study of model feedback in peer review~\cite{thakkar2026peerreview}, and asked reasoning models to find critical problems in papers~\cite{zhang2025reviewbench}. These studies judge what a review says about a paper as a whole. A fluent review can discuss a figure without comparing it with the text, so they do not show whether a figure was checked. We look inside each review for that specific act and record how far it gets.

\section{Study Design}
\label{sec:problem}

\paragraph{The task.}
To catch a figure--text contradiction in a complete manuscript, a reviewer has to pick the figure that deserves scrutiny, find the element and the claim that conflict, and check one against the other. A free-form review does not expose these steps separately. It shows only what the reviewer chose to write about the flawed figure. We therefore score each review by how far it gets, and we change one input at a time to see which changes carry reviews further. Each unit of the study is one full manuscript paired with one inconsistency documented in PRISMM-Bench~\cite{prismm}, which names a visual element (a figure or table) and the claim it conflicts with. The model receives the rendered pages and a neutral request to review the paper. It is not told that an error exists or where it is, and the documented inconsistency is used only afterwards to score the review. We enrolled 31 figure-bearing manuscripts whose PDFs and target elements could be rendered, using plans fixed before any review was read. They are longer than the benchmark's candidate pool (median 17 against 15 pages), and every claim below applies to this figure-centric set.

\paragraph{Four review outcomes.}
A review can stop at four points: it may omit the flawed element, mention it, raise a concern about it, or diagnose the documented contradiction. Let $L_1$ mark a review that names the target figure or table, $L_2$ one that names it in a sentence that also questions it, and $P$ one that two grader models both accept as identifying the documented contradiction with supporting quoted text. We report the nested rates $S_1 = L_1 \cup P$ (mention), $S_2 = L_2 \cup P$ (concern), and $S_3 = P$, which we call \emph{rubric-accepted diagnosis}. The rates describe what a review says. They do not separately measure candidate discovery, alignment, and verification, and a review can fail $S_3$ for reasons that we cannot assign to one step. Reviews that the graders cannot resolve (175 of 1{,}445) stay in the denominator as non-diagnoses.

\paragraph{Evaluation conditions.}
The \emph{open-ended cohort} contains 1{,}445 reviews: 31 papers, six models, two repeats, and four input conditions. In two conditions the model sees unmodified pages (the original and a byte-identical repeat, 722 reviews). In the other two, a border is drawn around the flawed figure or around an unrelated figure (723 reviews). No prompt names the figure or mentions an error, but two of the four conditions add a visual cue, so the pooled cohort is not cue-free. Diagnosis is 3.2\% on the unmodified pages and 3.3\% with borders, and Section~\ref{sec:null} analyzes the border contrast. The \emph{controlled arms} are a later run on unmodified PDFs with 186 reviews per arm (31 papers $\times$ six models). The unprompted arm uses the cohort's prompt and reaches a similar diagnosis rate (4.3\% against 3.3\%). The other arms add one ingredient of verification: the identity of the target figure (without saying that it contains an error), an instruction to check for contradictions, a crop of the flawed element placed in different locations, or the crop together with the conflicting clause. Two arms remove information. The text-only arm replaces the pages by the PDF text layer, with placeholders that name each figure but show nothing, and the OCR-only arm recovers the text printed inside figures without their rendered marks. The mention rate of the controlled arms is lower than that of the cohort (40.9\% against 60.6\%), because the cohort counts any reference to the registered element, including its caption or title, whereas the controlled arms count only figure and table identifiers. Applying that rule to the cohort's unmodified reviews gives 45.3\%. We therefore compare controlled arms only with each other, and the cohort only with itself.

\paragraph{Graders and what the audit supports.}
The six reviewers are Claude Sonnet~4.6, GPT-5.4, GPT-5~mini, GPT-5~nano, Gemini~3.1~Pro, and Gemini~3.1~Flash-Lite. Gemini~4~Argon and GPT-5.4 grade each review without seeing the model or condition, and agree at $\kappa = 0.62$ on 1{,}221 decidable pairs. A second pair of models replaying the rubric on 1{,}279 complete pairs accepts 2.74\% against 3.28\% for the primary pair. Two authors also audited 100 open-ended reviews, blind to model, condition, and grader label. The sample was stratified in advance and contains 20 of the 47 accepted diagnoses. The two authors agree at $\kappa = \HAkappaFlaw$ on whether a review diagnoses the documented flaw. Of the 19 decidable reviews that the rubric accepted, both authors confirmed only 10 (\HAprec\%), and no rejected review was confirmed by both. The rubric is therefore lenient on a sizable share of the reviews it accepts. This does not bound the population precision, because the sample is small, stratified, and drawn from the open-ended reviews, and the controlled arms have no human audit. A rubric-accepted rate is thus not a human-confirmed detection rate. A difference between arms is a difference in rubric-accepted diagnosis, and it reflects a difference in confirmed diagnosis only if rubric error is similar across arms. We did not test that, and the reference-free panel in the supplement is the only partial check on the controlled arms. Intervals are 95\% bootstrap intervals that resample papers, and paired contrasts use per-paper differences with an exact sign test. Table~\ref{tab:scope} lists what each analysis covers and how it was checked, and Supplementary Sections~\ref{app:scope} and~\ref{app:design} give selection, prompts, and per-model request settings.

\begin{table*}[t]
\centering\small
\setlength{\tabcolsep}{5pt}
\renewcommand{\arraystretch}{1.12}
\caption{\textbf{What each analysis covers.} All analyses use the same 31 manuscripts, except the paired excerpt, which needs a second concrete side to supply. The blind human audit sampled reviews from the open-ended cohort only. Controlled arms rely on the model rubric, with a reference-free panel as a second check (supplement).}\label{tab:scope}
\begin{tabular}{@{}p{3.6cm}p{1.4cm}p{4.9cm}p{5.2cm}@{}}
\toprule
Analysis & Section & Reviews and papers & Checked by\\
\midrule
Open-ended cohort & \ref{sec:null} & 1{,}445 reviews, 31 papers; four input conditions pooled & Rubric; two-model replay (1{,}279 pairs); blind audit (100 reviews)\\
Border highlights & \ref{sec:null} & 349 to 351 paired cells per contrast & Same as open-ended cohort\\
Controlled arms, $2 \times 2$ & \ref{sec:factorial} & 186 reviews per arm, 31 papers & Rubric; reference-free panel; no human audit\\
Crop placement, text-only, OCR-only & \ref{sec:factorial} & 186 reviews per arm, 31 papers & Same as controlled arms\\
Paired excerpt & \ref{sec:factorial} & 84 reviews, 14 papers & Same as controlled arms\\
Two-stage baseline & \ref{sec:twostage} & 186 runs, 31 papers & Same as controlled arms\\
\bottomrule
\end{tabular}
\end{table*}

\section{Results}
\label{sec:results}

Three questions take apart the single question of whether a model can verify visual evidence without being told what to verify. We first ask where unguided reviews stop (RQ1), then which single input change carries them further (RQ2), and finally whether the model can supply that input itself (RQ3). RQ1 draws on the open-ended cohort of 1{,}445 reviews; RQ2 and RQ3 draw on a later controlled run with 186 reviews per arm, and because the two runs count mentions differently we compare arms only within a run (Table~\ref{tab:scope}). Every rate is a rubric-accepted rate and not a human-confirmed detection rate (Section~\ref{sec:problem}).

\begin{figure*}[t]
\centering\includegraphics[width=\textwidth]{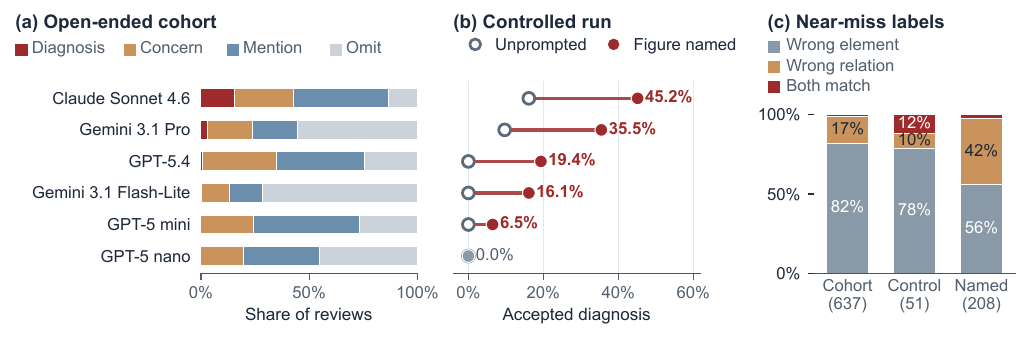}
\caption{\textbf{Reviews mention the flawed figure but rarely diagnose it, and naming the figure shifts the remaining misses toward the wrong relation.}
(a)~How far each model's open-ended reviews get in the cohort, ordered by rubric-accepted diagnosis.
(b)~Rubric-accepted diagnosis for the same models in the controlled run, unprompted and with the figure named (31 reviews per model and arm).
(c)~Grader label of near-miss records: open-ended cohort (637 grader records), controlled unprompted arm (51), and controlled arm with the figure named (208).}
\label{fig:reviewers}
\end{figure*}

\subsection{RQ1: Where Do Unguided Reviews Stop?}
\label{sec:null}
A review can fail by never reaching the flawed figure or by reaching it without checking it, and the four outcomes separate the two: the first leaves the figure unmentioned, the second mentions or questions it without diagnosing the contradiction. We trace the 1{,}445 open-ended reviews through the four outcomes, compare the flawed element with its nearest unflawed neighbor, and test whether a border around the figure changes what reviews say.

Of the 1{,}445 reviews, 39.4\% omit the flawed element, 34.2\% mention it without a concern, 23.2\% raise a concern that the graders do not accept, and 3.3\% (47 reviews) are accepted as diagnosing the documented contradiction (Figure~\ref{fig:reviewers}a). In nested terms, $S_1 = 60.6$\%, $S_2 = 26.4$\%, and $S_3 = 3.25$\% with a 95\% interval of [1.60, 5.18]. Diagnoses occur in 14 of the 31 manuscripts. Claude Sonnet~4.6 produces 37 of the 47, which is 15.3\% of its 242 reviews, while the other five models pool to 0.83\% [0.17, 1.65] although they mention the flawed element in 55.4\% of their reviews. One manuscript contributes 10 diagnoses, and leaving it out gives 2.65\%.

\paragraph{Mentions do not single out the flawed figure.}
Pairing each flawed element with its nearest unflawed neighbor of the same kind across 25 manuscripts gives 1{,}309 pairs. Reviews mention the flawed element at 57.9\% and the neighbor at 53.6\%, a difference of $+4.3$ points [$-5.5$, $+14.0$]. A visual highlight does not change this. A border around the flawed figure, compared with a border around an unrelated figure, shifts the concern rate by $+5.6$ points [$+0.8$, $+10.5$] and diagnosis by $+1.1$ points [$-0.4$, $+2.7$], and byte-identical repeats of the same reviews differ by a mean absolute 11.6 and 3.5 points per paper on the same two measures. Both shifts therefore lie within run-to-run variation. The borders enclose whole figures (median containment 0.999) and stay a median 49.5~pt from the caption, so they mark a figure but not the element or clause to compare.

\paragraph{Near-misses mostly lack the documented conflict.}
The graders label 82.1\% of 637 near-miss records as not matching the documented element, 16.8\% as matching the element but not the relation, and 1.1\% as matching both (Figure~\ref{fig:reviewers}c). The first label overstates misses of the figure itself: in the blind audit, at least one reader judged that the review critiques the target element in 21 of 30 sampled element-mismatch records. A reference-free check finds that most near-misses assert no cross-source conflict at all, 30.4\% of 276 against all 25 accepted reviews, and few reviews that miss the documented flaw state another genuine contradiction (\HAoffkey\% of audited element-mismatch records, frame-weighted).

Taken together, these outputs show that models do not simply skip the flawed figure. They mention it often, do not clearly favor it over its neighbors, and usually do not assert a conflict between sources, so a review that discusses a figure fluently is weak evidence that the figure was compared with the text. These are observations about review text. They do not show whether a model looked at the image, and a review can fail for reasons that the four outcomes cannot separate. Nor can the cohort say whether the same models would diagnose the contradiction if pointed to the figure, which RQ2 tests.

\subsection{RQ2: Which Interventions Improve Diagnosis?}
\label{sec:factorial}
RQ1 locates the gap between mentioning and diagnosing but not what would close it. A reviewer might lack the figure to check, the instruction to check, the second side of the comparison, or the rendered pixels. On the same 31 manuscripts and six models, we therefore add one such ingredient at a time (186 reviews per arm, unmodified PDFs) and compare each arm with the unprompted one. The controlled arms have no human audit, so a difference between arms is a difference in rubric-accepted diagnosis.

\paragraph{Naming the figure helps; asking for a check does not.}
Table~\ref{tab:factorial} crosses two changes: naming the target figure, without saying that it contains an error, and adding an instruction to check for contradictions. Unprompted review diagnoses \RAnoprompt\% \RAnopromptCI. Naming the figure raises this to \RAtarget\% \RAtargetCI, a gain of \RAtgtNoFrame{} points \RAtgtNoFrameCI. The consistency instruction alone adds \RAframeNoTarget{} points \RAframeNoTargetCI. With the figure named, it changes diagnosis by \RAframeTarget{} points \RAframeTargetCI, and the interaction is \RAinteract{} points \RAinteractCI. Every interval for the instruction includes zero. Naming the figure also raises the concern rate from 19.9\% to 82.3\%, and the mention rate to 100\% by construction, so the arm converts a search problem into a targeted one.

\begin{table}[t]
\centering\footnotesize
\setlength{\tabcolsep}{3.2pt}
\caption{\textbf{Naming the target figure and asking for a consistency check, crossed.} Each cell runs on the identical 31 manuscripts and six reviewers, with 186 reviews on the unmodified full PDF. $S_2$ is the concern rate and $S_3$ the rubric-accepted diagnosis rate, with paper-clustered 95\% CIs. Paired contrasts are per-paper differences in percentage points over $K = 31$.}\label{tab:factorial}
\begin{tabular}{@{}llrrrl@{}}
\toprule
Arm & Target & Verify & $S_2$ & $S_3$ & 95\% CI\\
\midrule
Unprompted PDF & No & No & 19.89 & 4.30 & [1.08, 8.06]\\
Global-consistency & No & Yes & 23.12 & 6.99 & [2.69, 12.37]\\
Target-pointed & Yes & No & 82.26 & 20.43 & [13.44, 28.49]\\
Target-consistency & Yes & Yes & 88.71 & 17.20 & [10.22, 25.27]\\
\midrule
\multicolumn{6}{@{}l@{}}{\textbf{Paired $S_3$ contrasts} (pp, $K = 31$)}\\
\multicolumn{3}{@{}l}{Target, no framing} & & $+16.13$ & [$+9.14$, $+24.19$]\\
\multicolumn{3}{@{}l}{Target, with framing} & & $+10.22$ & [$+4.30$, $+16.13$]\\
\multicolumn{3}{@{}l}{Framing, no target} & & $+2.69$ & [$-3.23$, $+9.14$]\\
\multicolumn{3}{@{}l}{Framing, with target} & & $-3.23$ & [$-10.22$, $+3.23$]\\
\multicolumn{3}{@{}l}{Interaction} & & $-5.91$ & [$-15.59$, $+2.15$]\\
\bottomrule
\end{tabular}

\end{table}

The gain is uneven across models (Figure~\ref{fig:reviewers}b). Claude Sonnet~4.6 rises from 16.1\% to 45.2\% and Gemini~3.1~Pro from 9.7\% to 35.5\%. GPT-5.4, Gemini~3.1~Flash-Lite, GPT-5~mini, and GPT-5~nano go from 0\% to 19.4\%, 16.1\%, 6.5\%, and 0\%. Pooled, the five models other than Sonnet rise from 1.9\% [0.0, 4.5] to 15.5\% [8.4, 23.2]. Per-model values rest on 31 reviews and are descriptive.

\paragraph{Naming the figure changes the character of the misses.}
Among near-miss grader records, the share that do not match the documented element falls from 78.4\% to 55.8\% once the figure is named, while the share that match the element but not the documented relation rises from 9.8\% to 41.8\% (Figure~\ref{fig:reviewers}c). Reviews with the figure named therefore often reach the right element and still state a different relation, and about four of five still do not diagnose the flaw. Naming the figure removes part of the search without settling how a specific element relates to the claim.

\paragraph{Supplying the conflicting clause adds no clear gain.}
If the missing ingredient were the second side of the comparison, handing it over should help. For the 14 manuscripts whose documented flaw has a concrete second side, we gave the model the crop of the flawed element beside the conflicting clause, so that neither the search nor the second side remained. Diagnosis is 15.5\% [4.8, 29.8], against 2.4\% [0.0, 6.0] unprompted and 22.6\% [10.7, 35.7] with the figure named alone on the same 14 manuscripts. The difference from naming the figure is $-7.1$ points [$-23.8$, $+10.7$], which is uninformative with 14 manuscripts. Of the 26 near-miss records in this arm, 20 (76.9\%) still critique something other than the documented element. Showing the right crop does not ensure that the review addresses it.

\paragraph{Coarse visual cues do not help.}
We also relocated the evidence rather than marking a whole figure as in RQ1. Placing the target crop beside the caption, in a matched band elsewhere on the page, or in the band that states the contradiction leaves diagnosis at 3.2\%, 1.6\%, and 3.2\%, all close to the unprompted 4.3\% and within the 3.5-point run-to-run swing. A reference-free panel that never saw the answer key gives a view not tied to the rubric. It finds more asserted contradictions with the figure named (22.6\% to 44.1\%) and no reliable change for the crop placements, yet also finds them in 36.0\% of text-only reviews, which diagnose 2.2\%. Asserting a conflict is thus not the same as diagnosing the documented one, and the consistency instruction raises the first without raising the second (Supplementary Section~\ref{app:assertion}).

\begin{figure*}[t]
\centering\includegraphics[width=\textwidth]{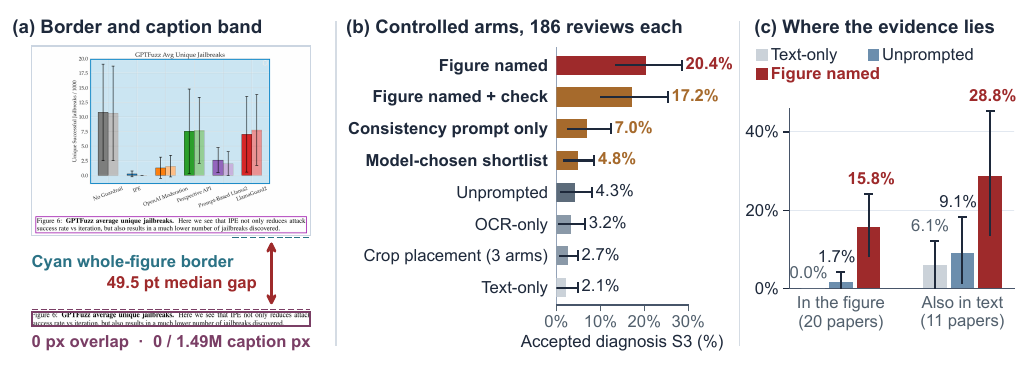}
\caption{\textbf{Naming the figure helps, while border highlights, crop placement, and a model-chosen shortlist leave diagnosis near the unprompted level.}
(a)~A second manuscript: the whole-figure border (cyan) lies a median 49.5~pt from the disputed caption band (magenta) and touches none of its pixels across 62 edits.
(b)~Rubric-accepted diagnosis for each arm of the controlled run (186 reviews, 31 manuscripts), with 95\% paper-clustered intervals.
(c)~Text-only, unprompted, and figure-named reviews by evidence type, on the 20 manuscripts whose evidence lies only in the figure and the 11 whose evidence is also stated in the text.}
\label{fig:probes}
\end{figure*}

\paragraph{Figure-only contradictions need the rendered figure.}
A reviewer with only the text layer can catch a contradiction whose evidence is also written in the text, but not one that exists only in the drawing. We split the manuscripts by each issue's own evidence statement, before any outcome was read. On the 20 manuscripts whose evidence lies only in the figure, such as an arrow route, a curve, or a bar height, text-only review diagnoses none of 120 reviews, unprompted review on rendered pages diagnoses 1.7\%, and naming the figure raises diagnosis to 15.8\% [8.3, 24.2] (Figure~\ref{fig:probes}c). On the 11 manuscripts whose evidence is also stated in the text, the corresponding values are 6.1\% for text-only review, 9.1\% unprompted, and 28.8\% [13.6, 45.5] with the figure named. This is what the task design predicts, and it shows that the rendered figure is needed to obtain any accepted diagnosis in that group. It does not show that the named-figure gain itself depends on the pixels, which would need a text-only arm with the figure named, and the stratum intervals are wide.

Among the ingredients we tested, naming the figure is the only one that raises rubric-accepted diagnosis substantially. It narrows the review to one figure, and the remaining misses less often involve the wrong element. A consistency instruction does not help, alone or with the figure named, which suggests that the reviewer lacks the object of the check and not the request to check. Borders and crop relocation mark where a figure is without saying which element or clause to compare, and adding the conflicting clause gives no clear gain beyond the name. Yet the named-figure arm still misses four of five contradictions, many by reaching the documented element and stating a different relation. No arm changes alignment without verification, so these experiments cannot say which of the two limits the models. What the named-figure arm supplies is exactly what a deployed reviewer lacks, which RQ3 examines.

\subsection{RQ3: Can Autonomous Candidate Selection Recover the Gain?}
\label{sec:twostage}
The RQ2 gain comes from the one input a deployed reviewer does not have: which figure to check. A reviewer has to produce it unaided, so we ask whether a model can, using a two-stage baseline that is a diagnostic of autonomy and not a proposed method. Stage~1 reads the unmodified PDF with no hint and ranks the five figures or tables most central to the paper's claims, with the caption claim and body claim of each, and is told not to report an error. Stage~2 returns the top three and their pages to the model with the consistency instruction in one review. The two stages use \RAtokenRatio{} times the tokens of a single pass, and a run with no usable identifier counts as a miss. We compare end-to-end diagnosis with the unprompted arm and with naming the figure, and we ask how often the shortlist contains the flawed figure.

Self-selected candidates leave diagnosis near the unprompted level. End-to-end diagnosis is \RAtwoStage\% \RAtwoStageCI. This is \RAtwoStageMinusBase{} points over unprompted review \RAtwoStageMinusBaseCI{}, which is consistent with no gain, and \RAtwoStageMinusTarget{} points below naming the figure \RAtwoStageMinusTargetCI. The baseline recovers \RArecovered\% of the gain from naming the figure, and the interval \RArecoveredCI{} is too wide to rule out either extreme.

One reason is visible in the shortlist. It contains the flawed figure in \RArecallThree\% of runs at rank three \RArecallThreeCI{} and in \RArecallFive\% at rank five \RArecallFiveCI, so the criterion we gave, centrality to the paper's claims, does not reliably surface the figure that contradicts its text. When the top three miss the flawed figure, diagnosis is 4.1\% [0.8, 8.3] over 122 runs. Candidate selection under this rule is therefore one clear reason the baseline falls short.

Selection is not the only limit, as an exploratory subset shows. When the shortlist does contain the flawed figure (64 runs, 23 papers), the batched check of three figures diagnoses \RAsTwoHitThree\% \RAsTwoHitThreeCI. On the same 64 runs, the single-figure arm with the figure named and the consistency instruction diagnoses 20.3\% [9.4, 31.4], and the paired difference is $-7.5$ points [$-13.9$, $-2.3$]. Against the single-figure arm with the figure named and no consistency instruction, which diagnoses 15.6\% [6.9, 24.2], the difference is $-1.5$ points [$-9.9$, $+9.7$], and the interval includes zero. The subset is defined by a Stage~1 outcome and was not planned, so the comparison is exploratory and compatible with both a loss from checking three figures at once and no loss.

A substantial gap therefore separates guided from autonomous verification: a model told where to look diagnoses roughly one contradiction in five, while a model that must decide where to look, ranking by centrality, stays at the unprompted level. The shortlist recall points to one cause, that centrality to a paper's claims is not the same as a figure's tendency to contradict its text. This result bounds one shortlist rule and no more. A shortlist ranked by the likelihood of a contradiction, or one that verifies each candidate in its own pass, is untested, so the finding does not show that autonomous selection is futile in general.

\section{Discussion}
\label{sec:discussion}

\paragraph{Verification starts with choosing what to verify.}
Checking visual evidence has a step before the check itself, which is deciding what to check. Existing benchmarks mostly measure the later step: the model receives a figure and a claim, or is told that something is wrong, and judges. Our manuscripts withhold all of that, and restoring one piece of it changes the outcome more than anything else we tried. Naming the figure raises rubric-accepted diagnosis well above the unprompted level, an instruction to check consistency does not, and a shortlist that the model builds by centrality stays far below the named-figure arm. Figure--text verification is therefore a selection problem joined to a judgment problem, and guided evaluations isolate only the judgment.

\paragraph{Implications for evaluation.}
Three consequences follow, each tied to a specific result. First, performance with a named target can overstate what a document assistant delivers on its own, since diagnosis with the figure named is about one in five while the model-chosen shortlist stays near the unprompted level. Second, the rendered figure has to be part of the test. For the 20 manuscripts whose contradiction exists only in the drawing, text-only review produced no accepted diagnosis, so a benchmark whose contradictions are also stated in prose cannot test visual verification. Third, a fluent review is not evidence of verification. Reviews mention the flawed figure about as often as a neighbor and usually assert no cross-source conflict, so scoring review quality or figure mentions can miss whether a figure was checked. An evaluation of this ability should hide the target, keep the pixels, and score the specific claim about the specific element.

\paragraph{Where the difficulty concentrates.}
Naming the figure shifts the difficulty and does not remove it. Once the figure is given, misses move from the wrong element toward the right element with the wrong relation, and adding the conflicting clause beside the crop gives no clear gain. What remains is the alignment of a visual element with the clause that bears on it, which prior work already finds weak on given figures~\cite{lal2025musciclaims,tong2024eyes}. Autonomous verification places that problem after a selection problem. Because our arms change inputs and not internals, they show where reviews stop and not why.

\paragraph{Scope.}
The findings rest on 31 figure-centric manuscripts that are longer than the benchmark's candidate pool, six models, and one model-based rubric that a blind audit found lenient. Absolute rates are probably high, and the controlled arms have no human audit, so a difference between arms reflects a difference in confirmed diagnosis only if rubric error is similar across arms. A reference-free panel agrees that naming the figure raises asserted conflicts and that relocating the crop does not, which is partial support. One model contributes most open-ended diagnoses, and per-model controlled values rest on 31 reviews. The paired-excerpt arm, the evidence-type split, and the shortlist-subset comparison are exploratory. The two-stage result bounds one selection rule. A rule that ranks figures by their tendency to contradict the text, or that verifies each candidate in its own pass, is the natural next test on this protocol.

\section{Conclusion}
Checking a figure against the text that cites it is a different ability from describing the figure, and autonomous verification is harder than guided verification. On 31 complete manuscripts with hidden, reviewer-documented inconsistencies, six frontier models mention the flawed figure in 60.6\% of 1{,}445 open-ended reviews, while a rubric accepts only 3.3\% as diagnosing the contradiction. In a separate controlled run, naming the figure raises rubric-accepted diagnosis from 4.3\% to 20.4\%, whereas a consistency instruction and coarse visual cues do not, and a shortlist the model builds by centrality reaches 4.8\%, far below the named-figure arm. Reliable multimodal reviewing therefore calls for models that discover which visual evidence to examine, align it with the claim it bears on, and verify the pair, and for evaluations that ask for all three without naming the target. The gap between mentioning a figure and diagnosing its contradiction is worth tracking as these systems improve.\label{lastbodypage}

\section*{LLM Usage Statement}
Six frontier multimodal foundation models, namely Claude Sonnet~4.6, Gemini~3.1~Pro, Gemini~3.1~Flash-Lite, GPT-5.4, GPT-5~mini, and GPT-5~nano, served as evaluated review subjects, while Gemini~4~Argon, GPT-5.4, GPT-6~Astra, and Claude~Opus~5 served as automated grading and source-anchored adjudication panels under locked protocols documented in Section~\ref{sec:problem} and Section~\ref{sec:results}. Large language model assistance supported analysis scripting, simulated reviewer feedback, and manuscript prose polishing. All empirical counts, statistical intervals, figures, tables, and scientific claims were deterministically verified against locked workspace logs by the authors, who take full responsibility for the manuscript.

\begingroup
\sloppy
\small
\renewcommand{\refname}{References}
\bibliographystyle{ieeenat_fullname}\bibliography{references}
\endgroup

\clearpage
\appendix
\renewcommand{\thesection}{\Alph{section}}
\numberwithin{table}{section}
\numberwithin{figure}{section}

\begin{center}
  {\large\bfseries\color{arxivblue} Supplementary Material}\\
  {\small\itshape\color{arxivgray} Reference tables and figures for the
   31-paper, 1{,}445-review cohort}
\end{center}
\noindent This supplement documents the 31-paper, 1{,}445-review cohort
behind the main-text results: the provider identifiers and call logs for
every model (Section~\ref{app:scope}), the review prompt and the cohort
selection procedure (Section~\ref{app:design}), the per-reviewer rates
(Section~\ref{app:reviewers}), the paper-level denominators
(Section~\ref{app:papers}), the reference-validity case study
(Section~\ref{app:reference}), the grading-rubric replay
(Section~\ref{app:grader}), the complete paired contrasts
(Section~\ref{app:contrasts}), the sibling control
(Section~\ref{app:visual}), the reference-anchored adjudication panel
(Section~\ref{app:anchor}), the factorial and two-stage baselines
(Section~\ref{app:factorial}), the human validation protocol
(Section~\ref{app:human}), the complete decoupled-probe results
(Section~\ref{app:probes}), the reference-free conflict check
(Section~\ref{app:assertion}), the claim-source taxonomy
(Section~\ref{app:taxonomy}), and the cohort robustness checks
(Section~\ref{app:moved}).

\begin{center}
\small
\setlength{\tabcolsep}{4.5pt}
\renewcommand{\arraystretch}{0.92}
\begin{tabular}{@{}lll@{}}
\toprule
Sec. & Contents & Floats\\
\midrule
A & Scope, provenance, and model identifiers & Tables A.1--A.3\\
B & Review prompt, cohort selection, and detectors & ---\\
C & Per-reviewer rates and leave-out sensitivity & Table C.1\\
D & Paper-level denominator audit and $31\times 6$ matrix & Table D.1\\
E & Reference validity case study & ---\\
F & Grading-rubric replay and agreement & Tables F.1--F.2\\
G & Complete paired intervention contrasts & Table G.1, Figure G.1\\
H & Sibling control, pixel audit, reference-free conflict & ---\\
I & Reference-anchored adjudication panel & Tables I.1--I.3\\
J & $2\times 2$ factorial and two-stage diagnostic baseline & Table J.1\\
K & Human validation protocol & ---\\
L & Complete decoupled-probe results & Table L.1\\
M & Reference-free conflict check across the probe arms & Table M.1\\
N & Claim-source taxonomy and pilot probes & Table N.1\\
O & Cohort robustness, grader checks, border-highlight contrasts
  & Tables O.1--O.3, Figure O.1\\
\bottomrule
\end{tabular}
\end{center}

\section{Scope, provenance, and model identifiers}
\label{app:scope}
This supplement documents the 31-paper, 1{,}445-review cohort analyzed in the main paper. All primary \texttt{P}/\texttt{N}/\texttt{U} review verdicts remain unchanged, and all cohort rates, paired contrasts, reference-free conflict adjudications, and replay statistics are deterministically reconstructed from archived artifacts using bootstrap seed \texttt{20261109} with 10{,}000 paper-clustered draws.

Every reviewer, grader, and adjudicator model is reported by the exact provider identifier recorded during execution (Table~\ref{tab:supp_models}), and Table~\ref{tab:supp_model_calls} lists the returned identifiers, request parameters, call counts, and execution windows read from the call logs. The cohort reviews were generated on \MVcohortGen{} and graded on \MVcohortGrade{}; the Phase~2 probes were generated on \MVprobeGen{} and graded on \MVprobeGrade{}; the adjudication panels ran from \MVadjud{}, all dates UTC. GPT-5.4 was requested under one identifier and the provider returned two strings, the dated snapshot \texttt{gpt-5.4-2026-03-05} and the bare alias \texttt{gpt-5.4}. In the cohort store \MVcohortDated{} of its calls returned the dated snapshot and \MVcohortAlias{} returned the alias; in Phase~2 \MVprobeDated{} returned the dated snapshot and \MVprobeAlias{} returned the alias, and the alias share of GPT-5.4 grader calls ranges from \MVgraderAliasMin\% to \MVgraderAliasMax\% across arms. The logs cannot resolve the alias to a snapshot, so the arms differ in the returned-string mix of one of the two graders, and we report a stratified check in place of a single version claim. Agreement between the two primary graders is \MVagreeDated\% on the \MVnDated{} Phase~2 reviews whose GPT-5.4 grader call returned the dated string and \MVagreeAlias\% on the \MVnAlias{} that returned the alias. Within the dated-string stratum, accepted diagnosis is \MVbaseDatedP/\MVbaseDatedN{} (\MVbaseDatedPct\%) for unprompted-full review and \MVtgtDatedP/\MVtgtDatedN{} (\MVtgtDatedPct\%) for target-pointed review; within the alias stratum it is \MVbaseAliasP/\MVbaseAliasN{} (\MVbaseAliasPct\%) and \MVtgtAliasP/\MVtgtAliasN{} (\MVtgtAliasPct\%). The strata are assigned by the provider and the comparison is descriptive and unpaired; a grader pinned to one snapshot remains a desirable refinement. Automated coding tools assisted with analysis scripts, source audits, figure generation, prose editing, and build verification; all statistics are computed directly from immutable archived outputs, and the authors take full responsibility for all content.

\begin{table*}[t]
\centering\footnotesize
\caption{\textbf{Provider model identifiers across roles in the audit.}}\label{tab:supp_models}
\begin{tabular}{@{}lp{0.70\textwidth}@{}}
\toprule
Role & Provider identifier(s)\\
\midrule
Frontier reviewers (six) & \texttt{claude-sonnet-4-6}, \texttt{gemini-3.1-pro-preview}, \texttt{gemini-3.1-flash-lite}, \texttt{gpt-5.4}, \texttt{gpt-5-mini}, \texttt{gpt-5-nano}\\
Primary rubric graders (two) & \texttt{gemini-4-argon}, \texttt{gpt-5.4}\\
Independent replay graders (two) & \texttt{gpt-6-astra}, \texttt{claude-opus-5}\\
Reference-free conflict adjudicator & \texttt{gpt-6-astra}\\
Exploratory decoupled-probe pilot & \texttt{gpt-5-mini-2025-08-07}\\
\bottomrule
\end{tabular}
\end{table*}

\begin{table*}[t]
\centering\scriptsize\setlength{\tabcolsep}{4pt}
\caption{\textbf{Returned API identifiers, request parameters, call counts, and execution windows by model and role.} Every cell is read from the local call logs. Calls are completed calls, shown against scheduled calls where the two differ. Two identifiers in one cell mean that the provider returned both strings for the same requested identifier.}\label{tab:supp_model_calls}
\resizebox{\textwidth}{!}{\begin{tabular}{@{}llp{3.7cm}lrl@{}}
\toprule
Model & Role & Returned API identifier(s) & Request parameters & Calls & Window (UTC) \\
\midrule
Claude Sonnet 4.6 & Reviewer, cohort & \texttt{claude-sonnet-4-6} & temp 0.7; max 16384 & 242/248 & 2026-11-03 \\
GPT-5.4 & Reviewer, cohort & \texttt{gpt-5.4}, \texttt{gpt-5.4-2026-03-05} & reasoning low; max 16384 & 241/248 & 2026-11-03 \\
GPT-5 mini & Reviewer, cohort & \texttt{gpt-5-mini-2025-08-07} & reasoning low; max 16384 & 243/248 & 2026-11-03 \\
GPT-5 nano & Reviewer, cohort & \texttt{gpt-5-nano-2025-08-07} & reasoning low; max 16384 & 239/248 & 2026-11-03 \\
Gemini 3.1 Pro Preview & Reviewer, cohort & \texttt{gemini-3.1-pro-preview} & temp 0.7; max 16384 & 241/248 & 2026-11-03 \\
Gemini 3.1 Flash Lite & Reviewer, cohort & \texttt{gemini-3.1-flash-lite} & temp 0.7; max 16384 & 239/248 & 2026-11-03 \\
\addlinespace[2pt]
Claude Sonnet 4.6 & Reviewer, Phase 2 & \texttt{claude-sonnet-4-6} & temp 0.7; max 16384/8192 & 355 & 2026-11-12 to 2026-11-15 \\
GPT-5.4 & Reviewer, Phase 2 & \texttt{gpt-5.4}, \texttt{gpt-5.4-2026-03-05} & reasoning low; max 16384/8192 & 355 & 2026-11-12 to 2026-11-15 \\
GPT-5 mini & Reviewer, Phase 2 & \texttt{gpt-5-mini}, \texttt{gpt-5-mini-2025-08-07} & reasoning low; max 16384/8192 & 355 & 2026-11-12 to 2026-11-15 \\
GPT-5 nano & Reviewer, Phase 2 & \texttt{gpt-5-nano}, \texttt{gpt-5-nano-2025-08-07} & reasoning low; max 16384/8192 & 354 & 2026-11-12 to 2026-11-15 \\
Gemini 3.1 Pro Preview & Reviewer, Phase 2 & \texttt{gemini-3.1-pro-preview} & temp 0.7; max 16384/8192 & 355 & 2026-11-12 to 2026-11-15 \\
Gemini 3.1 Flash Lite & Reviewer, Phase 2 & \texttt{gemini-3.1-flash-lite} & temp 0.7; max 16384/8192 & 355 & 2026-11-12 to 2026-11-15 \\
\addlinespace[2pt]
GPT-5.4 & Primary grader, cohort & \texttt{gpt-5.4-2026-03-05} & not logged & 1419/1451 & 2026-11-03 to 2026-11-04 \\
Gemini 4 Argon & Primary grader, cohort & \texttt{gemini-4-argon} & not logged & 1402/1451 & 2026-11-03 to 2026-11-04 \\
\addlinespace[2pt]
GPT-5.4 & Primary grader, Phase 2 & \texttt{gpt-5.4}, \texttt{gpt-5.4-2026-03-05} & reasoning low; max 2400 & 1943 & 2026-11-12 to 2026-11-15 \\
Gemini 4 Argon & Primary grader, Phase 2 & \texttt{gemini-4-argon} & temp 0.0; max 2400 & 1942/1943 & 2026-11-12 to 2026-11-15 \\
\addlinespace[2pt]
GPT-6 Astra & Panel-external replay & \texttt{gpt-6-astra}, \texttt{gpt-6-astra-2026-09-03} & not logged & 1452 & 2026-11-04 \\
Claude Opus 5 & Panel-external replay & \texttt{claude-opus-5} & not logged & 1422 & 2026-11-04 \\
\addlinespace[2pt]
GPT-6 Astra & Reference-free auditor, cohort & \texttt{gpt-6-astra} & not logged & 632 & 2026-11-06 \\
\addlinespace[2pt]
GPT-6 Astra & Reference-free auditor, Phase 2 & \texttt{gpt-6-astra} & temp 0.0; max 1200 & 1572 & 2026-11-12 \\
Claude Opus 5 & Reference-free auditor, Phase 2 & \texttt{claude-opus-5} & temp 0.0; max 1200 & 1572 & 2026-11-12 \\
\addlinespace[2pt]
Gemini 4 Argon & Source-anchored adjudicator & \texttt{gemini-4-argon} & temp 0.0; max 3000 & 44 & 2026-11-12 \\
GPT-6 Astra & Source-anchored adjudicator & \texttt{gpt-6-astra} & temp 0.0; max 3000 & 44 & 2026-11-12 \\
Claude Opus 5 & Source-anchored adjudicator & \texttt{claude-opus-5} & temp 0.0; max 3000 & 44 & 2026-11-12 \\
\bottomrule
\end{tabular}
}
\end{table*}

\begin{table*}[t]
\centering\footnotesize
\caption{\textbf{Selection funnel and representativeness of the full-PDF audit cohort.} Page counts exist for the \FUNpdfPapers{} release papers with an embedded source PDF; the remaining papers carry none. The final step is a manual, outcome-blind, batch-wise enrollment, and the number of non-enrolled candidates that were examined is not recorded in any local file.}\label{tab:suppfunnel}
\begin{tabular}{@{}lr@{\hspace{10pt}}r@{}}
\toprule
\multicolumn{3}{@{}l@{}}{\textbf{Selection funnel}}\\
Step & Inconsistencies & Papers\\
\midrule
Public PRISMM-Bench release & 384 & 353\\
Source PDF embedded in release & 204 & 191\\
Names figure or table (benchmark \texttt{category}) & 192 & 179\\
Names figure or table (parsed source) & 198 & 186\\
One issue per paper (candidate papers) & 186 & 186\\
Enrolled by the eight frozen plans & 31 & 31\\
\addlinespace[2pt]
\multicolumn{3}{@{}l@{}}{\textbf{Enrolled $K = 31$ cohort vs comparison populations}}\\
Feature & $K = 31$ & Comparison value\\
\midrule
Median pages & 17 & 15 (155 non-enrolled)\\
Mean pages & 19.0 & 16.1 (155 non-enrolled)\\
Pages IQR & $[15.5,\,20.0]$ & $[12,\,19]$ (155 non-enrolled)\\
Permutation $p$ (mean pages) & $0.009$ & \\
Figure-only issues & 29 & 112 (186 candidates)\\
Figure-only share (\%) & 93.5 & 60.2 (186 candidates)\\
Table-only issues & 0 & 70 (186 candidates)\\
Both figure and table & 2 & 4 (186 candidates)\\
Type-I / Type-II papers & 13 / 18 & \\
\bottomrule
\end{tabular}

\end{table*}

\section{Review prompt, cohort selection, and detectors}
\label{app:design}
All four visual conditions (unmodified \emph{Orig.}, byte-identical \emph{repeat}, \emph{target} figure highlight, and \emph{matched-random} figure highlight) use the identical open-ended peer-review prompt:
\begin{quote}\small
Write a thorough, critical peer review of the complete research paper provided in the page images, in order. Discuss its main contribution, strengths, methodological and experimental weaknesses, and questions for the authors. Ground specific criticisms in the paper and cite relevant sections, tables, or figures where useful. Distinguish demonstrated problems from uncertainty. Do not assume the paper contains an error. Write a normal prose review in English; no JSON is required.
\end{quote}
Each paper, reviewer model, and repeat index defines a paired matching cell; repeat indices do not share a fixed decoding seed.

The public PRISMM-Bench release registers \FUNrows{} inconsistencies over \FUNpapers{} papers. The source PDF ships with \FUNpdfRows{} of those issues over \FUNpdfPapers{} papers, \FUNvisRows{} issues over \FUNvisPapers{} papers name a figure or table under the parsed-source convention, and taking the first visual issue of each paper leaves \FUNcand{} candidate papers. The audit cohort is the \FUNenrolled{} papers enrolled from that candidate set by eight frozen execution plans, each frozen before its reviews were read, on technical grounds: source availability, a single renderable target element, and valid coordinates for the paired target and matched-random highlight rendering. Enrollment is manual, batch-wise, and independent of any reviewer outcome. \FUNnotenrolled{} candidates were not enrolled; the frozen plans record that other sources remained queued and name four explicit deferrals, one excluded misclassified title-legend pair, one real component-type mismatch, one gray-background region-encoding case, and one container-header mismatch, and no local record states a substantive exclusion rule, so none is claimed. Relative to the candidate set the cohort is longer, at \FUNmedPages{} median pages against \FUNmedPagesOther{} for the non-enrolled candidates, a difference of \FUNmeanPages{} against \FUNmeanPagesOther{} mean pages with a two-sided permutation $p = \FUNpermP$ over 10{,}000 permutations at seed 2027, and it is entirely figure-bearing: \FUNfigOnly{} figure-only and \FUNtableOnly{} table-only issues against \FUNfigOnlyOther{} and \FUNtableOnlyOther{} among the candidates. All claims in the main paper are scoped to this figure-centric full-PDF audit cohort. Table~\ref{tab:suppfunnel} gives the full comparison. Across $31\times 6\times 4\times 2 = 1{,}488$ planned slots, 1{,}445 returned review text (97.1\%); the 43 unreturned slots are distinct from the 175 returned reviews labeled \texttt{U} under the primary grading rubric.

Strict detector $L_1$ matches target figure or table identifiers; $L_2\subseteq L_1$ requires an identifier and a lexical concern or question marker in the same sentence (including broad tokens such as ``not'', ``unclear'', and question forms). Defining $S_1=L_1\cup P$, $S_2=L_2\cup P$, and $S_3=P$ yields strict totals of 870 ($L_1$) and 362 ($L_2$), nested totals of 876 ($S_1$) and 382 ($S_2$) with $|L_2\cap P|=27$, and four mutually exclusive states of 569 (no mention), 494 (mention only), 335 (concern only), and 47 (accepted diagnosis).

Dropping text/caption sources from the answer key to form a purely visual-only anchor yields 58.34\% ($S_1$), 24.91\% ($S_2$), and 3.25\% ($S_3$). Two papers carry secondary annotations (96 reviews; 15 annotation-level differences); alternative multi-issue aggregation rules yield $S_3$ rates of 2.91\%, 3.25\%, 3.25\%, and 3.46\%.

\section{Per-reviewer rates and leave-out sensitivity}
\label{app:reviewers}
Table~\ref{tab:supp_reviewers} reports per-reviewer nested rates on the full 31-paper cohort ($n=1{,}445$) and after excluding the disputed observation-space paper ($K=30$, $n=1{,}397$). Excluding the top-performing reviewer (\texttt{claude-sonnet-4-6}, 37/242 accepted diagnoses) across all 31 papers leaves $n=1{,}203$ reviews with $S_1=55.36\%$ $[47.26,63.18]$, $S_2=23.11\%$ $[16.51,30.61]$, and $S_3=0.83\%$ $[0.17,1.65]$ (10/1{,}203).
\sloppy
\emergencystretch=2em

\begin{table*}[t]
\centering\footnotesize
\caption{\textbf{Per-reviewer rates across all 31 papers and under leave-one-paper-out ($K=30$).} Intervals are 10{,}000-draw paper-clustered 95\% bootstrap CIs.}\label{tab:supp_reviewers}
\begin{tabular}{@{}lrrrrl@{}}
\toprule
Reviewer & $n$ & $S_1$ (\%) & $S_2$ (\%) & $S_3$ (\%) & $S_3$ 95\% CI\\
\midrule
\multicolumn{6}{@{}l@{}}{\emph{Primary cohort ($K=31$ papers, $n=1{,}445$ reviews)}}\\
\texttt{claude-sonnet-4-6} & 242 & 86.78 & 42.98 & 15.29 & [7.53, 24.37]\\
\texttt{gemini-3.1-pro-preview} & 241 & 44.40 & 23.65 & 2.90 & [0.41, 6.25]\\
\texttt{gpt-5.4} & 241 & 75.52 & 34.85 & 0.83 & [0.00, 2.51]\\
\texttt{gemini-3.1-flash-lite} & 239 & 28.45 & 12.97 & 0.42 & [0.00, 1.27]\\
\texttt{gpt-5-mini} & 243 & 73.25 & 24.28 & 0.00 & [0.00, 0.00]\\
\texttt{gpt-5-nano} & 239 & 54.81 & 19.67 & 0.00 & [0.00, 0.00]\\
\addlinespace[2pt]
\multicolumn{6}{@{}l@{}}{\emph{Leave-one-paper-out: excluding \texttt{8TyGCAuCGd} ($K=30$ papers, $n=1{,}397$ reviews)}}\\
\texttt{claude-sonnet-4-6} & 234 & 86.32 & 41.03 & 12.39 & [6.41, 19.33]\\
\texttt{gemini-3.1-pro-preview} & 233 & 43.78 & 23.61 & 2.15 & [0.00, 5.15]\\
\texttt{gpt-5.4} & 233 & 74.68 & 32.62 & 0.86 & [0.00, 2.60]\\
\texttt{gemini-3.1-flash-lite} & 231 & 28.14 & 12.99 & 0.43 & [0.00, 1.31]\\
\texttt{gpt-5-mini} & 235 & 72.34 & 21.70 & 0.00 & [0.00, 0.00]\\
\texttt{gpt-5-nano} & 231 & 54.11 & 19.91 & 0.00 & [0.00, 0.00]\\
\bottomrule
\end{tabular}
\end{table*}

\medskip
\noindent\textbf{Reading the per-reviewer rates.} Table~\ref{tab:supp_reviewers}
shows that the aggregate rate in the main paper is not an average over
interchangeable reviewers. The six reviewers differ by more than an order
of magnitude on the accepted endpoint: \texttt{claude-sonnet-4-6} reaches
$S_3 = 15.29\%$ while \texttt{gpt-5-mini} and \texttt{gpt-5-nano} never
produce an accepted diagnosis at all ($S_3 = 0.00\%$, with a degenerate
$[0.00, 0.00]$ interval). Two reviewers sit in between, and the ordering
on $S_1$ is not the ordering on $S_3$: \texttt{gpt-5.4} names the flawed
element often ($S_1 = 75.52\%$) but almost never converts the mention into
an accepted diagnosis ($S_3 = 0.83\%$), whereas
\texttt{gemini-3.1-pro-preview} names it less often ($S_1 = 44.40\%$) and
converts more often ($S_3 = 2.90\%$). The spread is therefore a spread in
\emph{assertion}, not in observation, and it is the reason the main paper
reports a pooled rate over reviewers instead of a per-reviewer one.
Removing the disputed observation-space paper barely moves any row, which
confirms that the disagreement between reviewers is a property of the
reviewers rather than of one difficult manuscript. Excluding the
top-performing reviewer entirely is the only change that moves the
aggregate, and it moves $S_3$ to $0.83\%$; the pooled $3.25\%$ is thus
carried by a minority of the reviewer pool, which is stated explicitly in
the main text and is the reason the human audit and the independent replay
in Table~\ref{tab:graders} are reported alongside it.

\section{Paper-level denominator audit and \texorpdfstring{$31\times 6$}{31x6} matrix}
\label{app:papers}
The tables below index all 31 papers in the cohort, pooling all four visual conditions and both repeats (with \texttt{U} retained in the denominator), followed by the $31\times 6$ accepted/returned matrix for reviewers (A)~\texttt{claude-sonnet-4-6}, (B)~\texttt{gemini-3.1-flash-lite}, (C)~\texttt{gemini-3.1-pro-preview}, (D)~\texttt{gpt-5-mini}, (E)~\texttt{gpt-5-nano}, and (F)~\texttt{gpt-5.4}.
\sloppy
\emergencystretch=2em

\begingroup\footnotesize\setlength{\tabcolsep}{5pt}\renewcommand{\arraystretch}{0.92}
\begin{longtable}{r l r r r @{\hspace{24pt}} r l *{6}{r}}
\caption{\textbf{Left: paper-level returned, accepted ($P$), and unresolved (\texttt{U}) counts. Right: accepted/returned counts by reviewer (A--F).}}\\
\toprule
ID & Paper ID & Ret. & $P$ & \texttt{U} & ID & Paper ID & (A) & (B) & (C) & (D) & (E) & (F)\\
\midrule\endfirsthead
\toprule
ID & Paper ID & Ret. & $P$ & \texttt{U} & ID & Paper ID & (A) & (B) & (C) & (D) & (E) & (F)\\
\midrule\endhead
01 & 8TyGCAuCGd & 48 & 10 & 4 & 01 & 8TyGCAuCGd & 8/8 & 0/8 & 2/8 & 0/8 & 0/8 & 0/8\\
02 & MazxSMs6Hs & 48 & 7 & 10 & 02 & MazxSMs6Hs & 6/8 & 0/8 & 1/8 & 0/8 & 0/8 & 0/8\\
03 & U9sHVjidYH & 48 & 6 & 10 & 03 & U9sHVjidYH & 3/8 & 0/8 & 3/8 & 0/8 & 0/8 & 0/8\\
04 & LXnTFMvn8A & 48 & 5 & 1 & 04 & LXnTFMvn8A & 2/8 & 1/8 & 0/8 & 0/8 & 0/8 & 2/8\\
05 & BeOEmnmyFu & 45 & 3 & 11 & 05 & BeOEmnmyFu & 3/8 & 0/7 & 0/8 & 0/7 & 0/7 & 0/8\\
06 & QO4bF6MHza & 43 & 3 & 1 & 06 & QO4bF6MHza & 3/8 & 0/7 & 0/7 & 0/7 & 0/7 & 0/7\\
07 & 0JwxMqKGxa & 48 & 2 & 3 & 07 & 0JwxMqKGxa & 2/8 & 0/8 & 0/8 & 0/8 & 0/8 & 0/8\\
08 & BpyHIrpUOL & 45 & 2 & 1 & 08 & BpyHIrpUOL & 2/7 & 0/7 & 0/8 & 0/8 & 0/7 & 0/8\\
09 & FNDudoox4A & 48 & 2 & 10 & 09 & FNDudoox4A & 2/8 & 0/8 & 0/8 & 0/8 & 0/8 & 0/8\\
10 & NFAjyyeSfB & 44 & 2 & 3 & 10 & NFAjyyeSfB & 2/7 & 0/6 & 0/7 & 0/8 & 0/8 & 0/8\\
11 & O2CG9B2k9Q & 47 & 2 & 3 & 11 & O2CG9B2k9Q & 2/8 & 0/8 & 0/8 & 0/8 & 0/7 & 0/8\\
12 & 3NMYMLL92j & 45 & 1 & 19 & 12 & 3NMYMLL92j & 0/7 & 0/8 & 1/8 & 0/8 & 0/7 & 0/7\\
13 & hgQFmoZT4g & 47 & 1 & 1 & 13 & hgQFmoZT4g & 1/8 & 0/7 & 0/8 & 0/8 & 0/8 & 0/8\\
14 & ttRSBZiCiu & 46 & 1 & 5 & 14 & ttRSBZiCiu & 1/8 & 0/8 & 0/7 & 0/7 & 0/8 & 0/8\\
15 & 0rXGGYNVAw & 47 & 0 & 1 & 15 & 0rXGGYNVAw & 0/8 & 0/8 & 0/8 & 0/8 & 0/7 & 0/8\\
16 & 3MDmM0rMPQ & 48 & 0 & 1 & 16 & 3MDmM0rMPQ & 0/8 & 0/8 & 0/8 & 0/8 & 0/8 & 0/8\\
17 & 8EaDOGMPUL & 45 & 0 & 3 & 17 & 8EaDOGMPUL & 0/8 & 0/8 & 0/8 & 0/8 & 0/7 & 0/8\\
18 & 8GMUa79ZKc & 45 & 0 & 9 & 18 & 8GMUa79ZKc & 0/7 & 0/8 & 0/8 & 0/7 & 0/7 & 0/8\\
19 & 9DDJuab67K & 47 & 0 & 5 & 19 & 9DDJuab67K & 0/8 & 0/8 & 0/7 & 0/8 & 0/8 & 0/8\\
20 & Aqfwhna1D7 & 48 & 0 & 21 & 20 & Aqfwhna1D7 & 0/8 & 0/8 & 0/8 & 0/8 & 0/8 & 0/8\\
21 & BrqFB8Nl7e & 47 & 0 & 4 & 21 & BrqFB8Nl7e & 0/8 & 0/8 & 0/8 & 0/8 & 0/8 & 0/7\\
22 & CfXRcN4iUw & 45 & 0 & 5 & 22 & CfXRcN4iUw & 0/8 & 0/6 & 0/8 & 0/8 & 0/8 & 0/7\\
23 & EvszuxrPTb & 47 & 0 & 0 & 23 & EvszuxrPTb & 0/8 & 0/8 & 0/8 & 0/8 & 0/7 & 0/8\\
24 & FLOaCQfZe9 & 48 & 0 & 7 & 24 & FLOaCQfZe9 & 0/8 & 0/8 & 0/8 & 0/8 & 0/8 & 0/8\\
25 & HCMmC8DETj & 47 & 0 & 1 & 25 & HCMmC8DETj & 0/7 & 0/8 & 0/8 & 0/8 & 0/8 & 0/8\\
26 & Mm3TSPRUKc & 47 & 0 & 1 & 26 & Mm3TSPRUKc & 0/8 & 0/8 & 0/8 & 0/8 & 0/7 & 0/8\\
27 & TppgZwyEnQ & 48 & 0 & 23 & 27 & TppgZwyEnQ & 0/8 & 0/8 & 0/8 & 0/8 & 0/8 & 0/8\\
28 & VL2txtmPYA & 48 & 0 & 3 & 28 & VL2txtmPYA & 0/8 & 0/8 & 0/8 & 0/8 & 0/8 & 0/8\\
29 & gKd8hH6ERY & 44 & 0 & 3 & 29 & gKd8hH6ERY & 0/8 & 0/7 & 0/7 & 0/8 & 0/8 & 0/6\\
30 & vfHISoWo2m & 46 & 0 & 3 & 30 & vfHISoWo2m & 0/7 & 0/8 & 0/8 & 0/8 & 0/8 & 0/7\\
31 & ywD00GsxgD & 48 & 0 & 3 & 31 & ywD00GsxgD & 0/8 & 0/8 & 0/8 & 0/8 & 0/8 & 0/8\\
\midrule
\multicolumn{2}{r}{Total} & 1{,}445 & 47 & 175 & \multicolumn{2}{r}{Total} & 37/242 & 1/239 & 7/241 & 0/243 & 0/239 & 2/241\\
\bottomrule
\end{longtable}
\endgroup

\section{Reference validity case study}
\label{app:reference}
Paper \texttt{8TyGCAuCGd} (\emph{Look Ma, No Training!}, ICLR 2024 OpenReview submission) contributes 10 of the 47 primary acceptances. Table~2 in that source counts Hill Climbing observation-space selections across 32 random initializations, whereas Figure~4a reports downstream trained RL return. Because the manuscript explicitly notes that the full-feature space achieves the highest baseline trained return while discussing search-cost trade-offs, a ranking difference between selection frequency and trained return does not by itself establish a literal contradiction, motivating our explicit $K=30$ leave-one-paper-out check.

\section{Grading-rubric replay and agreement}
\label{app:grader}
Primary dual-grader agreement is $\kappa=0.620$ across 1{,}221 mutually decidable pairs. The independent external replay (\texttt{gpt-6-astra} and \texttt{claude-opus-5}) issued 2{,}890 planned calls with 2{,}724 returned judgments, yielding 1{,}279 complete replay pairs (166 incomplete). On those 1{,}279 reviews, primary acceptance is 42/1{,}279 (3.28\%) versus 35/1{,}279 (2.74\%) on replay, giving an equal-paper paired difference of $-0.55$~pp $[-0.93,-0.23]$ (34 shared $P$, 8 primary-$P$ to replay-\texttt{U}, 1 primary-\texttt{U} to replay-$P$, 1{,}062 shared $N$, 26 primary-$N$ to replay-\texttt{U}, 118 primary-\texttt{U} to replay-$N$, and 30 shared \texttt{U}). Table~\ref{tab:supp_replay} breaks down three-way P/N/U $\kappa$ and $\Delta P$ across models and conditions, and Table~\ref{tab:supp_replay_matrix} gives the pooled confusion matrix behind these counts.

\begin{table}[t]
\centering\footnotesize
\setlength{\tabcolsep}{3.0pt}
\caption{\textbf{Primary vs.\ independent replay agreement across reviewers and visual conditions ($n=1{,}279$).}}\label{tab:supp_replay}
\begin{tabular}{@{}lrrrr@{}}
\toprule
Stratum & Complete & Missing & $\kappa$ & $\Delta P$ (pp)\\
\midrule
\texttt{claude-sonnet-4-6} & 214 & 28 & 0.591 & $-2.80$\\
\texttt{gemini-3.1-flash-lite} & 210 & 29 & 0.241 & $+0.00$\\
\texttt{gemini-3.1-pro-preview} & 207 & 34 & 0.318 & $-0.48$\\
\texttt{gpt-5-mini} & 214 & 29 & 0.012 & $+0.00$\\
\texttt{gpt-5-nano} & 220 & 19 & 0.294 & $+0.00$\\
\texttt{gpt-5.4} & 214 & 27 & 0.353 & $+0.00$\\
\midrule
\texttt{identical\_input} & 323 & 39 & 0.545 & $-0.31$\\
\texttt{matched\_random} & 321 & 42 & 0.359 & $+0.00$\\
\texttt{orig} & 317 & 43 & 0.478 & $-0.63$\\
\texttt{target} & 318 & 42 & 0.348 & $-1.26$\\
\bottomrule
\end{tabular}
\end{table}

\begin{table}[t]
\centering\footnotesize
\setlength{\tabcolsep}{4.2pt}
\caption{\textbf{Primary vs.\ independent replay label matrix ($n=1{,}279$ complete pairs).} Three-way P/N/U confusion matrix between the primary rubric and the independent replay; per-stratum $\kappa$ and $\Delta P$ are in Table~\ref{tab:supp_replay}.}\label{tab:supp_replay_matrix}
\begin{tabular}{@{}lrrrr@{}}
\toprule
 & \multicolumn{3}{c}{\textbf{Independent replay label}} & \\
\cmidrule(lr){2-4}
Primary label & Replay P & Replay N & Replay U & Total\\
\midrule
Primary P & 34 & 0 & 8 & 42\\
Primary N & 0 & 1{,}062 & 26 & 1{,}088\\
Primary U & 1 & 118 & 30 & 149\\
\midrule
Total & 35 & 1{,}180 & 64 & 1{,}279\\
\bottomrule
\end{tabular}
\end{table}

\section{Complete paired intervention contrasts}
\label{app:contrasts}
\begin{figure*}[t]
\centering\includegraphics[width=\textwidth]{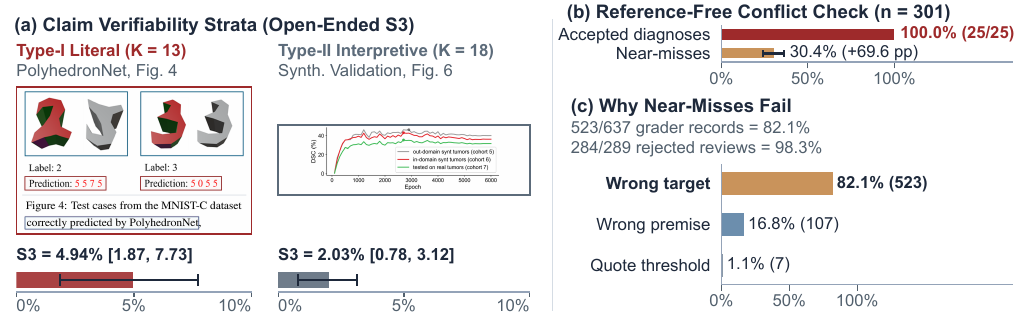}
\caption{\textbf{Claim verifiability strata and reference-free near-miss grounding breakdown across the open-ended cohort.}
(a) Authentic high-resolution source crops contrasting Type-I literal mismatches ($K = 13$, $S_3 = 4.94\%$) against Type-II interpretive disputes ($K = 18$, $S_3 = 2.03\%$).
(b) Reference-free conflict check across 301 order-stable concerns (100.0\% accepted diagnoses vs.\ 30.4\% near-misses).
(c) Failure stage breakdown across 637 near-miss grader records and 289 rejected concern reviews.}
\label{fig:audit}
\end{figure*}

Table~\ref{tab:supp_contrasts} reports both nested ($S_1, S_2, S_3$) and strict ($L_1, L_2, P$) complete-pair contrasts across all 31 papers and under $K=30$ exclusion. Across the 12 primary nested contrasts, minimum Holm--Bonferroni adjusted $p_{\text{Holm}}=0.767$ ($0.575$ across the 9 intervention contrasts). Figure~\ref{fig:audit} decomposes the same cohort along a different axis, separating contradictions whose evidence is literal from those whose evidence is interpretive, and breaking down where reference-free conflict checks and near-miss grading fail.

\begin{table*}[t]
\centering\footnotesize
\setlength{\tabcolsep}{4.2pt}
\caption{\textbf{Paired visual-intervention contrasts under nested and strict definitions for $K=31$ papers (left) and $K=30$ excluding \texttt{8TyGCAuCGd} (right).}}\label{tab:supp_contrasts}
\begin{tabular}{@{}lllrrlr @{\hspace{16pt}} rrlr@{}}
\toprule
& & & \multicolumn{4}{c}{$K=31$ papers} & \multicolumn{4}{c}{$K=30$ papers}\\
\cmidrule(lr){4-7}\cmidrule(l){8-11}
Def. & Contrast & Endpt. & Cells & Diff. & 95\% CI & $p$ & Cells & Diff. & 95\% CI & $p$\\
\midrule
nested & Target ${-}$ Orig. & $S_1$ & 349 & $-1.88$ & $[-6.09, +2.57]$ & 0.210 & 337 & $-1.67$ & $[-6.00, +2.85]$ & 0.286\\
nested & Target ${-}$ Orig. & $S_2$ & 349 & $+3.84$ & $[-0.10, +7.67]$ & 0.263 & 337 & $+3.13$ & $[-0.75, +6.88]$ & 0.359\\
nested & Target ${-}$ Orig. & $S_3$ & 349 & $+0.87$ & $[-0.37, +2.49]$ & 0.625 & 337 & $+0.62$ & $[-0.61, +2.22]$ & 1.000\\
nested & Target ${-}$ Random & $S_1$ & 351 & $+1.19$ & $[-2.99, +5.25]$ & 0.286 & 339 & $+1.23$ & $[-3.00, +5.33]$ & 0.286\\
nested & Target ${-}$ Random & $S_2$ & 351 & $+5.64$ & $[+0.83, +10.50]$ & 0.064 & 339 & $+4.99$ & $[+0.17, +9.90]$ & 0.093\\
nested & Target ${-}$ Random & $S_3$ & 351 & $+1.11$ & $[-0.43, +2.71]$ & 0.289 & 339 & $+0.87$ & $[-0.69, +2.44]$ & 0.453\\
nested & Target ${-}$ Repeat & $S_1$ & 350 & $-0.92$ & $[-5.54, +3.71]$ & 0.845 & 338 & $-0.67$ & $[-5.46, +4.09]$ & 1.000\\
nested & Target ${-}$ Repeat & $S_2$ & 350 & $+4.19$ & $[+0.10, +8.20]$ & 0.134 & 338 & $+3.77$ & $[-0.32, +7.79]$ & 0.189\\
nested & Target ${-}$ Repeat & $S_3$ & 350 & $+0.65$ & $[-2.15, +3.49]$ & 1.000 & 338 & $+0.67$ & $[-2.22, +3.58]$ & 1.000\\
nested & Orig.\ ${-}$ Repeat & $S_1$ & 350 & $+1.39$ & $[-3.65, +6.47]$ & 0.678 & 338 & $+1.44$ & $[-3.75, +6.64]$ & 0.678\\
nested & Orig.\ ${-}$ Repeat & $S_2$ & 350 & $+0.44$ & $[-4.89, +5.57]$ & 0.690 & 338 & $+0.74$ & $[-4.74, +5.99]$ & 0.541\\
nested & Orig.\ ${-}$ Repeat & $S_3$ & 350 & $-0.22$ & $[-2.64, +2.25]$ & 1.000 & 338 & $+0.05$ & $[-2.47, +2.55]$ & 1.000\\
\midrule
strict & Target ${-}$ Orig. & $L_1$ & 349 & $-2.48$ & $[-6.81, +2.02]$ & 0.134 & 337 & $-2.28$ & $[-6.73, +2.37]$ & 0.189\\
strict & Target ${-}$ Orig. & $L_2$ & 349 & $+3.00$ & $[-0.77, +6.68]$ & 0.383 & 337 & $+2.55$ & $[-1.32, +6.31]$ & 0.503\\
strict & Target ${-}$ Random & $L_1$ & 351 & $+0.63$ & $[-3.44, +4.64]$ & 0.503 & 339 & $+0.65$ & $[-3.49, +4.72]$ & 0.503\\
strict & Target ${-}$ Random & $L_2$ & 351 & $+4.80$ & $[+0.27, +9.52]$ & 0.152 & 339 & $+4.40$ & $[-0.19, +9.25]$ & 0.210\\
strict & Target ${-}$ Repeat & $L_1$ & 350 & $-1.51$ & $[-6.18, +3.31]$ & 0.701 & 338 & $-1.29$ & $[-6.11, +3.63]$ & 0.845\\
strict & Target ${-}$ Repeat & $L_2$ & 350 & $+3.60$ & $[-0.46, +7.65]$ & 0.115 & 338 & $+3.44$ & $[-0.86, +7.58]$ & 0.167\\
strict & Orig.\ ${-}$ Repeat & $L_1$ & 350 & $+1.39$ & $[-3.65, +6.47]$ & 0.678 & 338 & $+1.44$ & $[-3.75, +6.64]$ & 0.678\\
strict & Orig.\ ${-}$ Repeat & $L_2$ & 350 & $+0.69$ & $[-4.83, +5.92]$ & 0.424 & 338 & $+0.99$ & $[-4.57, +6.34]$ & 0.307\\
\bottomrule
\end{tabular}
\end{table*}

\section{Sibling control, pixel audit, and reference-free conflict check}
\label{app:visual}
\textbf{Matched sibling control.} Pairing each target figure or table with the nearest-ordinal non-issue element of the same type yields 1{,}309 pairs across 1{,}167 reviews and 25 papers (6 papers have no eligible sibling). Equal-paper lexical mention is 57.91\% (target) vs.\ 53.59\% (sibling; $+4.33$~pp $[-5.46,+13.98]$), and lexical concern is 21.92\% vs.\ 18.90\% ($+3.01$~pp $[-4.31,+10.17]$).

\textbf{Pixel-level edit localization.} Across all 62 executed edits (31 target changing 796{,}094 pixels; 31 matched-random changing 697{,}465 pixels; 1{,}493{,}559 total), 0 altered pixels enter any PDF caption band (0/31 target captions; 0/30 caption-bearing control pages; closest target approach 12.5~pt, median 49.5~pt). In PANDA, the legend highlight changes 14{,}571 pixels (mean RGB delta 13.0/255); in MathHay, the header highlight changes 21{,}359 pixels (mean RGB delta 25.7/255) while leaving the mislabeled caption untouched.

\textbf{Reference-free conflict check.} Of 382 nested concerns ($S_2$), 316 decidable lexical-concern reviews ($27\ L_2\cap P$ and $289\ L_2\cap N$) were evaluated in two page-order-swapped passes by \texttt{gpt-6-astra} with benchmark answer keys, visual conditions, and grader labels withheld (95.3\% agreement; 15 unstable excluded; 301 order-stable retained, including 1 unresolved counted as non-affirmative; 66 concerns [$20\ P\setminus L_2$ and $46\ L_2\cap U$] were not adjudicated). Among the 301 stable items, 109 (36.2\%, 95\% CI $[30.0,41.9]$) assert a cross-source contradiction (25/25 accepted vs.\ 84/276 [30.4\%] near-miss; $+69.6$~pp $[63.1,76.8]$ over 30 papers), corresponding to 28.5\%--32.7\% across adjudicated order-uncertainty on all 382 concerns (50.0\% worst-case Manski upper bound).

\section{Reference-anchored adjudication panel}
\label{app:anchor}
The cohort's $S_3$ label rests on a dual-LLM rubric, and Section~\ref{app:grader} bounds that rubric with a panel-external replay. The replay re-reads the same review under the same rubric, so it cannot say whether the rubric's reference answers are themselves the right reading of the paper. We therefore ran an independent \emph{reference-anchored} panel: three models that are neither the cohort's grading pair nor the replay panel (\texttt{claude-opus-5}, \texttt{gemini-4-argon}, \texttt{gpt-6-astra}), each given the review, the registered issue (the paper's own reference answer), and the source pages, and asked to adjudicate whether the review raises that relation. The panel is a third opinion, not a restatement of either existing instrument.

The sampling frame is the cohort's own 44-item human audit packet (23 papers, four strata: \emph{discovery\_paper}, \emph{disputed\_discovery}, \emph{near\_miss}, \emph{object\_correct}). All 44 items were adjudicated with 0 panel-member failures, so coverage is complete. Table~\ref{tab:supp_anchor} reports the panel against the primary rubric.

\begin{table}[t]
\centering\footnotesize
\setlength{\tabcolsep}{4.5pt}
\caption{\textbf{Reference-anchored panel vs.\ the primary rubric on the 44-item packet.} Rows are the primary label, columns the panel's majority label (a tied panel is counted \texttt{U}).}\label{tab:supp_anchor}
\begin{tabular}{@{}lrrr@{}}
\toprule
 & \multicolumn{3}{c}{\textbf{Panel label}} \\
\cmidrule(lr){2-4}
Primary label & P & N & U\\
\midrule
P & 6 & 0 & 3\\
N & 1 & 28 & 0\\
U & 0 & 5 & 1\\
\midrule
Total & 7 & 33 & 4\\
\bottomrule
\end{tabular}
\end{table}

Three-way agreement with the primary rubric is 79.5\% with $\kappa = 0.556$. On the decidable $P$-versus-not binary ($n = 35$, excluding items the panel or the rubric labels \texttt{U}), the panel recovers all 6 primary acceptances (recall 1.00) with one false positive (precision 0.857), giving $\kappa = 0.906$. The 9 disagreements are not noise: 5 are primary-\texttt{U} items that the reference-anchored panel resolves to \texttt{N}, which is expected because the rubric's \texttt{U} is emitted whenever the two graders disagree or a contract check fails, whereas the anchor panel reads against the paper directly. The 3 primary-\texttt{P} items that the panel labels \texttt{U} fall in the \emph{disputed\_discovery} stratum (agreement 66.7\%, against 90.0\% for \emph{near\_miss} and 85.7\% for \emph{discovery\_paper}), which is exactly the stratum where the benchmark's own reference answer is contestable. Eight items (H005, H014, H015, H017, H018, H031, H032, H036) were flagged reference-ambiguous by at least one panel member; they are the items a human pass must resolve first.

This three-model panel provides an independent source-anchored adjudication across all 44 cohort audit items, quantifying instrument distance against direct reading of the manuscript pages and complementing the 96-item exposed-label human source-consistency check conducted on the three-paper development pilot (95/96 = 98.96\% agreement across two 48-item batches). For orientation, Table~\ref{tab:suppfacts} collects every arm reported in the main paper with its review count and rubric-accepted diagnosis rate, so the panel and replay results can be read against a single table of arm-level outcomes.

\begin{table}[t]
\centering\footnotesize
\caption{\textbf{Arms reported by the main paper.} Every arm runs on all 31 manuscripts and up to six reviewers with the unmodified full PDF unless stated; $n$ is the number of graded reviews, and rubric-accepted diagnosis $S_3$ carries a paper-clustered 95\% CI. Full per-arm results are in Table~\ref{tab:supp_probes} and Table~\ref{tab:factorial}.}\label{tab:suppfacts}
\begin{tabular}{@{}lrr@{}}
\toprule
Arm & $n$ & $S_3$ (\%)\\
\midrule
Unprompted full PDF & 186 & 4.30\\
Target-pointed & 186 & 20.43\\
Target-consistency & 186 & 17.20\\
Global-consistency & 186 & see Table~\ref{tab:factorial}\\
Two-stage index-then-verify & 186 & see Table~\ref{tab:supptwostage}\\
Crop placement, three arms & 558 & 2.69\\
OCR-only & 186 & 3.23\\
Text-only & 186 & 2.15\\
\bottomrule
\end{tabular}
\end{table}

\section{\texorpdfstring{$2\times 2$}{2x2} factorial and two-stage diagnostic baseline}
\label{app:factorial}
The main-paper factorial runs four arms on the identical 31 manuscripts, six reviewers, and 186 reviews per arm, all on the unmodified full PDF and all scored by the unchanged dual-grader rubric. \emph{unprompted-full} is the cohort's neutral prompt with neither factor. \emph{global-consistency} adds one sentence: every figure and table is checked against its caption and against the rest of the paper, and any contradiction is reported concretely, with nothing named. \emph{target-pointed} names the registered element and its page without asserting that a defect exists. \emph{target-consistency} names the element and asks whether it is consistent with the rest of the paper, which is the arm described as \emph{target-consistency} throughout. Cells, intervals, and all paired contrasts are in Table~\ref{tab:factorial}.

The two-stage pipeline uses two model calls per paper and reviewer. Stage one receives the unmodified full PDF and no hint, and returns the five figures or tables most central to the paper's claims with the caption claim and the body claim of each, then ranks three pairs for verification; it is explicitly told not to assume an error and not to report one. The Stage~1 instruction is quoted verbatim below. Stage two receives the same full PDF with the target-consistency instruction, in which the model's own three ranked identifiers and pages replace the oracle element; the review returned by stage two is the unified audit report that reaches the grader. No benchmark metadata enters either stage. The benchmark key is used afterward, for recall of the registered element in the Stage~1 shortlist and for grading the Stage~2 review. A cell whose Stage~1 answer yields no usable identifier has no Stage~2 call and stays in the denominator as a non-diagnosis, so the end-to-end rate is an intention-to-treat estimate over all 186 cells.

\begin{quote}\small
You are given a complete research paper as page images, in order. Build a claim-figure index for this paper. Step 1: select the 5 figures or tables that are most central to the paper's claims. Step 2: for each one, record its identifier exactly as printed in the paper, such as Figure 3 or Table 2, the number of the page image on which it appears, the specific quantitative or structural claim that its caption makes about it, and the specific quantitative or structural claim that the body text makes about it. Quote the paper's wording where possible. Step 3: rank the 3 figure-claim pairs for which checking the rendered figure or table against the stated claims matters most for the paper's conclusions. Do not assume the paper contains an error, and do not report any error at this step. Return only a JSON object with the keys candidates and verify\_top3, whose candidates each record element, page, caption\_claim, and body\_claim, and whose verify\_top3 lists three identifiers.
\end{quote}

Table~\ref{tab:supptwostage} reports the Stage~1 recall, the end-to-end diagnosis rate, the share of the oracle gain recovered, and the token cost, all computed from the logged runs.

\begin{table}[t]
\centering\footnotesize
\caption{\textbf{Two-stage diagnostic baseline in aggregate.} Recall is the share of 186 paper-model cells whose Stage~1 shortlist contains the registered element; end-to-end $S_3$ counts a cell without a usable identifier as a non-diagnosis. Token counts are as logged.}\label{tab:supptwostage}
\begin{tabular}{@{}lrl@{}}
\toprule
Two-stage quantity & Value & 95\% CI\\
\midrule
Stage 1 target recall@3 & 34.41\% & [23.66, 46.24]\\
Stage 1 target recall@5 & 57.53\% & [43.55, 71.51]\\
End-to-end $S_3$ ($n = 186$) & 4.84\% & [1.61, 8.60]\\
$S_3$ gain over unprompted & $+0.54$ pp & [$-3.76$, $+5.38$]\\
Oracle gain recovered & 3.3\% & [$-33.3$, $+27.3$]\\
Tokens vs.\ single pass & 2.04$\times$ & \\
\bottomrule
\end{tabular}

\end{table}

\paragraph{Shortlist-subset comparison.}
The main paper reports an exploratory comparison on the 64 runs whose Stage~1 top-three shortlist contains a target element (23 papers). It was computed from the logged per-run grades with the loaders, estimator, and seed of the main analysis (\texttt{scripts\_v2/routeA\_subset\_check.py}, read-only). On these runs the two-stage review diagnoses 4 of 64 (6.25\%, 95\% CI [0.0, 13.85]). The single-figure target-consistency arm diagnoses 13 of 64 (20.31\%, [9.38, 31.43]), a paired two-stage difference of $-7.54$ points [$-13.91$, $-2.32$] with sign test $p = 0.031$ (0 positive, 6 negative of 6 papers with a difference). The target-pointed arm diagnoses 10 of 64 (15.62\%, [6.90, 24.18]), a difference of $-1.52$ points [$-9.86$, $+9.71$] with $p = 0.125$ (1 positive, 6 negative of 7). The unprompted arm diagnoses 4 of 64 (6.25\%, [0.0, 13.21]), a difference of $+3.19$ points [$-5.22$, $+14.20$]. The subset is defined by a Stage~1 outcome, so it is not a randomized comparison, and the conclusion depends on which single-figure arm is the reference.

\section{Human validation protocol}
\label{app:human}
The dual-author audit draws \HAn{} reviews from the 1{,}445-review cohort in strata fixed before annotation: 20 rubric-accepted diagnoses, 30 wrong-target near-misses with the conflict-asserting half oversampled 20 to 10, 20 wrong-premise near-misses including all five that both graders ground, 15 unresolved items, and 15 uncritical mentions or omissions as controls. Two annotators receive independently shuffled CSVs stripped of every model name, visual arm, repeat index, and automated grader verdict; identifiers are a seeded permutation, so a row's position carries no stratum signal. The hidden key stores each item's true stratum with its cell frame size, so oversampling can be undone by frame weighting. Each annotator marks whether the review critiques the target element, whether it verifies the documented flaw, and whether it states a different figure-text contradiction that the annotator confirmed in the source PDF. Scoring reports Cohen's $\kappa$ between annotators on each judgment, agreement of their consensus with the primary rubric with precision and recall, and the share of near-misses carrying a valid off-key contradiction, with paper-clustered bootstrap intervals at 10{,}000 draws and seed 2027. The scoring script refuses to emit any statistic while either sheet is incomplete or holds an invalid code, so no partial annotation can be reported as a result.

\section{Complete decoupled-probe results}
\label{app:probes}
Table~\ref{tab:supp_probes} reports every Phase~2 probe arm with its full
bootstrap interval and its paired contrast against the in-experiment
baseline. The \emph{unprompted-full} arm reproduces the cohort's own neutral
prompt on the rendered pages under the same rubric and is the baseline for
every contrast. The \emph{target-pointed} arm adds a pointer to the target
element without asserting that a flaw exists; \emph{target-consistency} keeps the
pointer and asks explicitly whether the two sources are consistent; the
\emph{paired-excerpt} arm supplies the element crop and the conflicting text
clause side by side, removing search from the task; the \emph{text-only}
and \emph{OCR-only} arms remove or restrict the image channel; and the three
\emph{crop-placement} arms re-render the target crop into the caption band, into a
matched band elsewhere on the page, and into the band in which the
contradiction is actually stated.
Rates are percentages with 95\% paper-clustered bootstrap CIs (10{,}000
draws, \texttt{seed=20261109}); contrasts are per-paper paired differences
over the papers the two arms share, with an exact per-paper sign test. Every
contrast must be read against its own metric-matched identical-input band
($\pm 10.57$, $\pm 11.56$, and $\pm 3.54$~pp for $S_1$, $S_2$, and $S_3$),
never against another metric's band.

An arm is reported as measured only when every one of its reviews carries
two grader verdicts that both returned a usable result. A review whose
grader calls failed at the transport layer carries no verdict, and is
re-graded rather than counted as a non-acceptance; had such records been
counted as rejections, a grading outage would have been reported as a model
result. All nine arms in Table~\ref{tab:supp_probes} are measured, over
$1{,}564$ reviews in total, and every reported rate is computed over
reviews for which both graders returned a verdict.

Two grader-verdict contract failures survived the first grading pass and
were repaired before the numbers were computed. In both cases the grader
had transcribed a span of the review correctly but had dropped the
review's markdown emphasis markers, so the span was a literal quote of the
review's words but not a literal substring of the review's raw text. The
substring check now compares the two with emphasis markers removed from
both sides; because the removal is applied to the review as well, a span
that is not in the review still fails to match. The change can only widen
what counts as a literal span, so it can repair a verdict that was rejected
for a presentational difference but cannot alter the label of a verdict
that already passed.

\begin{table*}[t]
\centering\footnotesize
\setlength{\tabcolsep}{4.0pt}
\caption{\textbf{Complete decoupled-probe results with bootstrap intervals and exact sign tests.}}
\label{tab:supp_probes}
\begin{tabular}{llrrrlrlr}
\toprule
& & & & \multicolumn{2}{c}{\emph{Rate (\%)}} & \multicolumn{2}{c}{\emph{$\Delta$ vs.\ baseline (pp)}} & \\
\cmidrule(lr){5-6}\cmidrule(lr){7-8}
Arm & Metric & $n$ & $K$ & Rate & 95\% CI & $\Delta$ & 95\% CI & $p$ \\
\midrule
Unprompted full (baseline) & $S_1$ & 186 & 31 & 40.86 & [32.80, 48.92] & -- & -- & -- \\
 & $S_2$ &  &  & 19.89 & [13.98, 26.34] & -- & -- & -- \\
 & $S_3$ &  &  & 4.30 & [1.08, 8.06] & -- & -- & -- \\
\addlinespace[1.5pt]
Target-pointed & $S_1$ & 186 & 31 & 100.00 & [100.00, 100.00] & $+59.14$ & [$+51.08$, $+67.20$] & $<0.001$ \\
 & $S_2$ &  &  & 82.26 & [76.88, 87.10] & $+62.37$ & [$+53.76$, $+70.43$] & $<0.001$ \\
 & $S_3$ &  &  & 20.43 & [13.44, 28.49] & $+16.13$ & [$+9.14$, $+24.19$] & $<0.001$ \\
\addlinespace[1.5pt]
Consistency-prompt & $S_1$ & 186 & 31 & 100.00 & [100.00, 100.00] & $+59.14$ & [$+51.08$, $+67.20$] & $<0.001$ \\
 & $S_2$ &  &  & 88.71 & [84.41, 93.01] & $+68.82$ & [$+60.75$, $+76.34$] & $<0.001$ \\
 & $S_3$ &  &  & 17.20 & [10.22, 25.27] & $+12.90$ & [$+5.91$, $+20.43$] & $0.003$ \\
\addlinespace[1.5pt]
Paired excerpt & $S_1$ & 84 & 14 & 71.43 & [59.52, 82.14] & $+36.90$ & [$+21.43$, $+52.38$] & $0.003$ \\
 & $S_2$ &  &  & 30.95 & [19.05, 44.05] & $+15.48$ & [$+2.38$, $+28.57$] & $0.109$ \\
 & $S_3$ &  &  & 15.48 & [4.76, 29.76] & $+13.10$ & [$+3.57$, $+25.00$] & $0.062$ \\
\addlinespace[1.5pt]
Text-only & $S_1$ & 186 & 31 & 43.01 & [31.72, 54.84] & $+2.15$ & [$-6.99$, $+11.29$] & $0.678$ \\
 & $S_2$ &  &  & 16.13 & [9.14, 24.19] & $-3.76$ & [$-10.22$, $+2.69$] & $0.189$ \\
 & $S_3$ &  &  & 2.15 & [0.00, 4.84] & $-2.15$ & [$-5.38$, $+0.54$] & $0.375$ \\
\addlinespace[1.5pt]
OCR-only & $S_1$ & 186 & 31 & 39.78 & [29.57, 50.54] & $-1.08$ & [$-8.60$, $+6.45$] & $0.839$ \\
 & $S_2$ &  &  & 15.59 & [9.68, 22.58] & $-4.30$ & [$-10.22$, $+1.08$] & $0.383$ \\
 & $S_3$ &  &  & 3.23 & [0.54, 6.45] & $-1.08$ & [$-3.76$, $+1.61$] & $0.688$ \\
\addlinespace[1.5pt]
Caption-band placement & $S_1$ & 186 & 31 & 33.87 & [25.27, 42.47] & $-6.99$ & [$-15.59$, $+1.61$] & $0.167$ \\
 & $S_2$ &  &  & 12.90 & [8.06, 18.28] & $-6.99$ & [$-13.98$, $+0.00$] & $0.093$ \\
 & $S_3$ &  &  & 3.23 & [0.54, 6.45] & $-1.08$ & [$-4.84$, $+2.69$] & $0.688$ \\
\addlinespace[1.5pt]
Matched-band placement & $S_1$ & 186 & 31 & 37.63 & [30.11, 45.70] & $-3.23$ & [$-11.29$, $+4.84$] & $0.832$ \\
 & $S_2$ &  &  & 15.59 & [10.22, 21.51] & $-4.30$ & [$-10.75$, $+2.15$] & $0.481$ \\
 & $S_3$ &  &  & 1.61 & [0.00, 3.76] & $-2.69$ & [$-5.91$, $+0.00$] & $0.219$ \\
\addlinespace[1.5pt]
Disputed-prose placement & $S_1$ & 186 & 31 & 37.63 & [29.57, 45.70] & $-3.23$ & [$-11.83$, $+5.38$] & $0.405$ \\
 & $S_2$ &  &  & 19.89 & [13.98, 26.34] & $+0.00$ & [$-5.38$, $+5.38$] & $1.000$ \\
 & $S_3$ &  &  & 3.23 & [1.08, 6.45] & $-1.08$ & [$-5.38$, $+3.23$] & $1.000$ \\
\addlinespace[1.5pt]
\bottomrule
\end{tabular}

\end{table*}

The rubric instantiation used for the \emph{paired-excerpt} arm is a
documented, frozen deviation: the cohort rubric is written for a complete
ordinary peer review, and the paired-excerpt arm produces a focused
consistency verdict on one supplied pair, so the rubric's description of
its input is re-instantiated for that arm. The relation-matching criteria,
the P/N/U decision, the evidence-quote requirement, the confidence
threshold, and the consensus rule are unchanged, so a P under the
instantiated rubric means exactly what a P under the cohort rubric means.

\section{Reference-free conflict check across the probe arms}
\label{app:assertion}

The rubric used to grade the cohort scores whether a review
\emph{identifies} the documented flaw. A second, independent
measurement asks a different question: does the review explicitly
\emph{assert} that two manuscript elements contradict each other?
The two are not interchangeable. A review may name a figure, voice a
generic concern about it, and still never make the relational claim
at all, so a rubric that counts \emph{concerns} cannot by itself
establish that a model failed to perform the comparison. Each review
in every arm was therefore put to a two-member reference-free panel
(\texttt{claude-opus-5} and \texttt{gpt-6-astra}) that read the
review alone with benchmark answer keys, arm metadata, and grader verdicts withheld, and decided whether it asserts a cross-source
contradiction; the panel never saw the rubric label, the arm, or
the other panel member's verdict. A record counts as affirmative only
when both members returned a usable verdict and a majority asserted
a contradiction; records in which a panel member failed at the
transport layer were re-adjudicated rather than counted as negative
votes.

All 1,572 adjudicated requests returned a usable
verdict from both members (0 records required a retry), so the panel
is complete on every arm. Table~\ref{tab:supp_assertion} reports the
affirmative rate with a paper-clustered bootstrap interval and the
paired per-paper difference against the unprompted-full baseline, using
the same estimator and seed as the rest of this supplement.

\textbf{The panel reproduces the rubric's ordering on a different
quantity.} Under the rubric, pointing at the target lifts $S_3$ from
4.30\% to 20.43\% and asking for a consistency check
lifts it to 17.20\%. Under the panel, the same two
interventions
raise the assertion rate from 22.58\% to
44.09\% (+21.51~pp) and to
53.76\% (+31.18~pp).
Removing search entirely with the paired-excerpt arm lands at
38.10\%, between the two. Both measurements show that
naming the figure raises the rate and that relocating the crop does
not. They differ on the consistency instruction, which raises asserted
conflicts by more than naming the figure alone but does not raise
rubric-accepted diagnosis, so asserting a conflict and diagnosing the
documented one are different quantities.

\textbf{Relocating the evidence changes nothing.} The three crop-placement
arms move the target crop to a different band of the same page
without changing the prompt. Under the panel, re-rendering the crop
into its own caption band gives 27.42\% and
into the band in which the contradiction is stated gives
22.04\% ($-0.54$~pp,
$p = 1.000$); a matched band
elsewhere on the page gives 23.12\%
(+0.54~pp,
$p = 0.791$). None of the three
separates from the baseline, and none separates the caption band
from a band placed elsewhere on the page. Placing the evidence where
the contradiction is written is therefore not sufficient to make a
reviewer assert the comparison.

\begin{table*}[t]
\centering\footnotesize
\setlength{\tabcolsep}{4.0pt}
\caption{\textbf{Reference-free conflict check: share of reviews in
which the panel affirms that two manuscript elements contradict
each other, with paper-clustered bootstrap intervals and exact sign
tests against the unprompted-full baseline.}}
\label{tab:supp_assertion}
\begin{tabular}{lrrrlrr}
\toprule
Arm & $n$ & $K$ & Affirmative rate & 95\% CI & $\Delta$ vs.\ baseline (pp) & $p$ \\
\midrule
Unprompted full (baseline) & 186 & 31 & 22.58 \& [18.82, 26.34] & -- & -- \\
\addlinespace[1.5pt]
Target-pointed & 186 & 31 & 44.09 \& [37.63, 50.00] & $+21.51$ \& [$+13.98$, $+29.03$] & $<0.001$ \\
\addlinespace[1.5pt]
Consistency-prompt & 186 & 31 & 53.76 \& [48.39, 59.68] & $+31.18$ \& [$+24.19$, $+38.71$] & $<0.001$ \\
\addlinespace[1.5pt]
Paired excerpt & 84 & 14 & 38.10 \& [23.81, 53.57] & $+16.67$ \& [$+1.19$, $+33.33$] & $0.227$ \\
\addlinespace[1.5pt]
Text-only & 186 & 31 & 36.02 \& [31.18, 40.86] & $+13.44$ \& [$+8.06$, $+18.82$] & $<0.001$ \\
\addlinespace[1.5pt]
OCR-only & 186 & 31 & 33.33 \& [28.49, 38.17] & $+10.75$ \& [$+5.38$, $+16.13$] & $0.001$ \\
\addlinespace[1.5pt]
Caption-band placement & 186 & 31 & 27.42 \& [23.66, 31.18] & $+4.84$ \& [$-0.54$, $+10.22$] & $0.210$ \\
\addlinespace[1.5pt]
Matched-band placement & 186 & 31 & 23.12 \& [19.89, 26.34] & $+0.54$ \& [$-3.76$, $+4.84$] & $0.791$ \\
\addlinespace[1.5pt]
Disputed-prose placement & 186 & 31 & 22.04 \& [18.82, 25.81] & $-0.54$ \& [$-5.38$, $+3.76$] & $1.000$ \\
\addlinespace[1.5pt]
\bottomrule
\end{tabular}

\end{table*}

\section{Claim-source taxonomy and pilot probes}
\label{app:taxonomy}Classifying the 31 documented issues prior to viewing reviews separates $K=13$ Type-I literal mismatches (30/607 accepted, 4.94\% $[1.87,7.73]$) from $K=18$ Type-II interpretive disputes (17/838 accepted, 2.03\% $[0.78,3.12]$), an exploratory gap of $+2.91$~pp (10{,}000-draw paper-clustered 95\% CI $[-0.82,+7.00]$). Rule~1 (claim source is \texttt{expectation}) and Rule~2 (either claim or evidence source is \texttt{expectation}) induce the identical $13/18$ split ($+2.91$~pp) because every issue's evidence field in this cohort cites a concrete manuscript element; Rule~3 (reclassifying normative claims with numeric/lexical literals as Type-I, $23/8$ split) yields $+2.20$~pp (3.82\% vs.\ 1.62\%).

\textbf{MathHay lexical check \& eight-call decoupled probe.} All three accepted MathHay reviews (\texttt{QO4bF6MHza}) quote the erroneous caption string, name MSMD, and cite Section~3.2 without repeating ``Multi-Step'' verbatim inside the critique sentence. In an exploratory eight-call pilot (\texttt{gpt-5-mini-2025-08-07}, 4{,}096-token cap) across four progressive cueing stages (full paper, target page, paired crops, paired crops plus relational prompt) on MathHay and PANDA, only the relational prompt elicits the MathHay task conflict, while PANDA remains unsolved even when cued.

\begingroup\footnotesize\setlength{\tabcolsep}{5pt}\renewcommand{\arraystretch}{0.94}
\begin{longtable}{l l l l r r}
\caption{\textbf{Complete 31-paper claim-source taxonomy.} Type-I issues cite an explicit manuscript claim source; Type-II issues cite an implicit domain \texttt{expectation}.}\label{tab:taxonomy}\\
\toprule
Paper ID & Stratum & Claim source & Evidence source & $n$ & $P$\\
\midrule\endfirsthead
\multicolumn{6}{@{}l}{\small\itshape Table \thetable{} continued from previous page}\\
\toprule
Paper ID & Stratum & Claim source & Evidence source & $n$ & $P$\\
\midrule\endhead
8TyGCAuCGd & Type-I & table\_2 & figure\_4a & 48 & 10\\
MazxSMs6Hs & Type-I & caption & plot & 48 & 7\\
LXnTFMvn8A & Type-I & text & Figure 1 & 48 & 5\\
QO4bF6MHza & Type-I & caption & figure\_17 & 43 & 3\\
BpyHIrpUOL & Type-I & caption & figure\_4 & 45 & 2\\
FNDudoox4A & Type-I & caption & Figure 2 & 48 & 2\\
ttRSBZiCiu & Type-I & figure\_1(b) & algorithm\_1 & 46 & 1\\
0rXGGYNVAw & Type-I & theorem & figure\_2(a) & 47 & 0\\
8EaDOGMPUL & Type-I & caption & Figure 4 & 45 & 0\\
9DDJuab67K & Type-I & text & Figure 1(a) & 47 & 0\\
BrqFB8Nl7e & Type-I & text & Figure 2 & 47 & 0\\
EvszuxrPTb & Type-I & figure\_1a & figure\_2a & 47 & 0\\
TppgZwyEnQ & Type-I & caption & Figure 3 & 48 & 0\\
\midrule
U9sHVjidYH & Type-II & expectation & Figure 5 & 48 & 6\\
BeOEmnmyFu & Type-II & expectation & figure\_8 & 45 & 3\\
0JwxMqKGxa & Type-II & expectation & Figure 1 & 48 & 2\\
NFAjyyeSfB & Type-II & expectation & figure\_1b, equation\_5 & 44 & 2\\
O2CG9B2k9Q & Type-II & expectation & Figure 4 and Figure 5 & 47 & 2\\
3NMYMLL92j & Type-II & expectation & Figure 1 & 45 & 1\\
hgQFmoZT4g & Type-II & expectation & Figure 1 caption & 47 & 1\\
3MDmM0rMPQ & Type-II & expectation & figure\_6 & 48 & 0\\
8GMUa79ZKc & Type-II & expectation & (3) and figure\_3 & 45 & 0\\
Aqfwhna1D7 & Type-II & expectation & figure\_3 & 48 & 0\\
CfXRcN4iUw & Type-II & expectation & Figure 2(c) and Table 1 & 45 & 0\\
FLOaCQfZe9 & Type-II & expectation & Figure 2 & 48 & 0\\
HCMmC8DETj & Type-II & expectation & figure\_2 & 47 & 0\\
Mm3TSPRUKc & Type-II & expectation & figure\_4 & 47 & 0\\
VL2txtmPYA & Type-II & expectation & figure\_6 & 48 & 0\\
gKd8hH6ERY & Type-II & expectation & figure\_3 & 44 & 0\\
vfHISoWo2m & Type-II & expectation & figure\_a & 46 & 0\\
ywD00GsxgD & Type-II & expectation & figure\_6 & 48 & 0\\
\bottomrule
\end{longtable}
\endgroup

\medskip
\noindent\textbf{Reading the taxonomy.} Table~\ref{tab:taxonomy} lists all
31 documented issues with the claim source and the evidence source that the
review cited, together with the review count $n$ and the number $P$ of
reviews the rubric accepted. Two patterns are visible. First, the accepted
reviews concentrate in the top of the list: the four largest $P$ values
(10, 7, 6, and 5) all belong to Type-I issues whose claim source is a
caption or a table cell, which is the stratum where the mismatch is
literal and the evidence pointer is unambiguous. Second, the claim source
varies far more than the evidence source. Every evidence entry names a
concrete manuscript element --- a figure, a caption, an algorithm, an
equation --- while the claim source is \texttt{expectation} for all 18
Type-II issues, which is why Rules~1 and~2 in the main text induce the same
$13/18$ split: there is no case in this cohort where the evidence source
alone is an expectation. The rows also show where the ceiling on $P$ comes
from. With $n \approx 45$ reviews per paper, a paper whose issue is
reliably named can reach $P \approx 10$ (8TyGCAuCGd) but never more,
because the rubric requires the diagnosis to survive both graders and the
contract checks. Fifteen of the 31 papers have $P = 0$, and every one of
them is a paper where the claim source is a sentence or a theorem rather
than a caption or a table cell: the models notice that something is off in
the prose but cannot connect it to the figure, which is the failure mode
the main paper's target-consistency arm is designed to remove.

\section{Cohort robustness, grader checks, and border-highlight contrasts}
\label{app:moved}
\begin{figure}[!t]
\centering\includegraphics[width=0.97\textwidth]{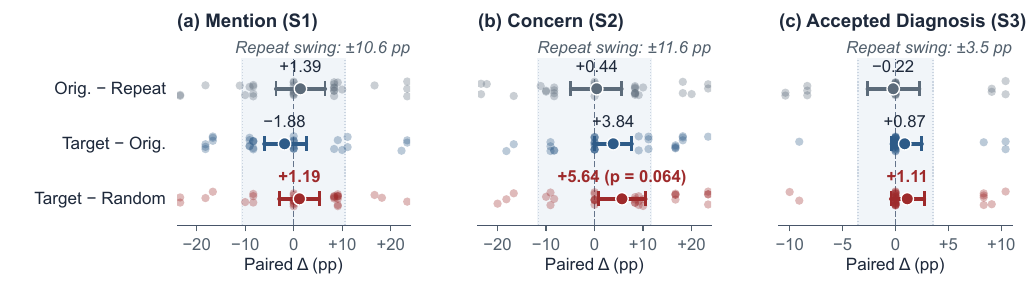}
\caption{\textbf{Whole-figure highlights produce small shifts within natural per-paper repeat variability and leave rubric-accepted diagnosis unchanged.}
Panels a to c plot paired paper-level contrasts on $S_1$ mention, $S_2$ concern, and $S_3$ rubric-accepted diagnosis for Orig.\ minus Repeat, Target minus Orig., and Target minus Random across $K = 31$ papers, with shaded empirical repeat variability bands of $\pm 10.6$, $\pm 11.6$, and $\pm 3.5$~pp.}
\label{fig:contrasts}
\end{figure}

This section collects the checks on the open-ended cohort that the main-text
results on the open-ended cohort summarize. It answers three questions that
the main text raises but does not pursue. First, how much of the headline
rate depends on a particular paper, a particular reviewer, or a particular
anchor definition of the rubric (Table~\ref{tab:funnel}). Second, whether
the rubric's accepted diagnoses would survive an instrument that never saw
the cohort, a source-anchored panel, a third grader, and a blind human pass
(Table~\ref{tab:graders}). Third, whether the border-highlight interventions
that the main text uses to make the figure visible move the measurement
more than the model's own run-to-run noise does (Table~\ref{tab:visual} and
Figure~\ref{fig:contrasts}). Every number below is computed from the same
archived artifacts as the main text, with the same 10{,}000-draw
paper-clustered bootstrap.

\noindent\textbf{Reading the cohort exclusions.} Every row of Table~\ref{tab:funnel}
reaches the same qualitative conclusion as the main text: the rubric-accepted
diagnosis rate stays near three percent. The primary cohort gives
$S_3 = 47/1{,}445 = 3.25\%$. Dropping the single most influential paper
(leave-one-paper-out, $K = 30$) moves the estimate to $2.65\%$, so no single
manuscript carries the result. Dropping the reviewer whose reviews overlap
with the grader shortlists moves it further, to $0.83\%$, which is the
strictest of the three exclusions and shows that the headline rate is
sensitive to reviewer composition. Re-anchoring the rubric on the visual
channel alone, with no answer-key text, leaves $S_3$ unchanged at $3.25\%$
while $S_1$ falls from $60.62\%$ to $58.34\%$; the models name the flawed
figure almost as often without being told that an answer key exists, but
they still rarely convert that mention into an accepted diagnosis. The
stratum split is the largest effect in the table: Type-I literal mismatches
are accepted at $4.94\%$ against $2.03\%$ for Type-II interpretive disputes,
and the two strata differ by $+2.91$ points with a 95\% CI of
$[-0.82, +7.00]$ that includes zero, so the split is suggestive rather than
established.

\begin{table}[!t]
\centering\footnotesize
\setlength{\tabcolsep}{4.0pt}
\renewcommand{\arraystretch}{1.06}
\hyphenpenalty=10000 \exhyphenpenalty=10000
\caption{\textbf{Rubric-accepted diagnosis stays near three percent across cohort exclusions, anchor definitions, and verifiability strata.} The cohort pools four input conditions, including figure borders. All $1{,}445$ returned reviews remain in $n$, including 175 unresolved \texttt{U} labels. Intervals are $10{,}000$-draw paper-clustered bootstrap 95\% CIs over $K$ paper clusters.}\label{tab:funnel}
\begin{tabular}{@{}lrrl@{}}
\toprule
Population / criterion & Count & Rate (\%) & 95\% CI\\
\midrule
\multicolumn{4}{@{}l@{}}{\textbf{Primary cohort} ($K = 31$ papers, $n = 1{,}445$ reviews)}\\
$S_1 = L_1 \cup P$ & 876 & 60.62 & {$[53.76,\ 67.36]$}\\
$S_2 = L_2 \cup P$ & 382 & 26.44 & {$[19.75,\ 34.07]$}\\
$S_3 = P$ & 47 & 3.25 & {$[1.60,\ 5.18]$}\\
\addlinespace[2pt]
\multicolumn{4}{@{}l@{}}{\textbf{Leave-one-paper-out} ($K = 30$ papers, $n = 1{,}397$ reviews)}\\
$S_1 = L_1 \cup P$ & 838 & 59.99 & {$[52.82,\ 66.74]$}\\
$S_2 = L_2 \cup P$ & 354 & 25.34 & {$[18.69,\ 32.86]$}\\
$S_3 = P$ & 37 & 2.65 & {$[1.35,\ 4.20]$}\\
\addlinespace[2pt]
\multicolumn{4}{@{}l@{}}{\textbf{Leave-top-reviewer-out} ($K = 31$ papers, $n = 1{,}203$; excl.\ Sonnet 4.6)}\\
$S_1 = L_1 \cup P$ & 666 & 55.36 & {$[47.26,\ 63.18]$}\\
$S_2 = L_2 \cup P$ & 278 & 23.11 & {$[16.51,\ 30.61]$}\\
$S_3 = P$ & 10 & 0.83 & {$[0.17,\ 1.65]$}\\
\addlinespace[2pt]
\multicolumn{4}{@{}l@{}}{\textbf{Visual-only anchor} ($K = 31$ papers, $n = 1{,}445$; no answer-key text)}\\
$S_1 = L_1 \cup P$ & 843 & 58.34 & {$[51.60,\ 64.84]$}\\
$S_2 = L_2 \cup P$ & 360 & 24.91 & {$[18.58,\ 32.39]$}\\
$S_3 = P$ & 47 & 3.25 & {$[1.60,\ 5.18]$}\\
\addlinespace[2pt]
\multicolumn{4}{@{}l@{}}{\textbf{By issue stratum} ($S_3$ accepted diagnosis)}\\
Type-I literal ($K = 13$, $n = 607$) & 30 & 4.94 & {$[1.87,\ 7.73]$}\\
Type-II interp.\ ($K = 18$, $n = 838$) & 17 & 2.03 & {$[0.78,\ 3.12]$}\\
\bottomrule
\end{tabular}

\end{table}

\noindent\textbf{Reading the grader checks.} The rubric's own numbers are
stable across every independent instrument in Table~\ref{tab:graders}. The
panel-external replay, run by two models that never graded the cohort,
returns $2.74\%$ against the primary rubric's $3.28\%$, an equal-paper
paired difference of $-0.55$~pp $[-0.93, -0.23]$. The source-anchored
44-item panel agrees with the rubric on $79.55\%$ of items with three-way
$\kappa = 0.556$, and reaches $\kappa = 0.906$ on the decidable
$P$-versus-not binary. The upper bound that counts every unresolved review
as accepted gives $15.36\%$, roughly five times the primary rate, which
bounds how much of the residual uncertainty could be closed by resolving
\texttt{U} labels alone. The dual-author blind audit of 100 reviews
reports $\kappa = 0.898$ for flaw detection against the rubric and
$\HAagree\%$ agreement, with precision $\HAprec\%$ at recall $\HArec\%$: the
audit confirms about half of the rubric's acceptances, and every
rubric-accepted review it marked wrong was a wrong-target mistake rather
than a spurious flaw. The off-key rate is $\HAoffkey\%$, so the rubric
almost never flags a figure that is not the registered issue.

\begin{table}[!t]
\centering\footnotesize
\setlength{\tabcolsep}{2.2pt}
\renewcommand{\arraystretch}{1.03}
\hyphenpenalty=10000 \exhyphenpenalty=10000
\caption{\textbf{Observed diagnosis rates hold between 2.74 and 3.74 percent across grading panels, whereas counting every unresolved review as accepted gives 15.36 percent. The human audit, which covers the open-ended cohort only, confirmed about half of the rubric-accepted reviews it sampled.} Top: primary rubric, panel-external two-model replay over $n = 1{,}279$ pairs, source-anchored 44-item adjudication over items \texttt{H001} to \texttt{H044}, pilot source check, and assumption-based bounds. Bottom: replay matrix.}\label{tab:graders}
\begin{tabular}{@{}llr@{}}
\toprule
Protocol / audit & Panel or rule & Outcome\\
\midrule
\multicolumn{3}{@{}l@{}}{\textbf{Primary rubric vs.\ independent replay} ($n = 1{,}279$)}\\
Primary rubric & Gemini 4 Argon, GPT-5.4 & 42 (3.28\%)\\
External replay & GPT-6 Astra, Claude Opus 5 & 35 (2.74\%)\\
Paired diff. & within paper & $-0.55$\;{[$-0.93$, $-0.23$]}\\
\addlinespace[2pt]
\multicolumn{3}{@{}l@{}}{\textbf{Source-anchored 44-item audit} (Opus 5, Argon, Astra)}\\
3-way agreement & $\kappa = 0.556$ & 35/44 (79.55\%)\\
Binary P vs.\ N & $\kappa = 0.906$, 100.0\% rec. & 34/35 (97.14\%)\\
Divergences (9) & U$\to$N / P$\to$U / N$\to$P & 5 / 3 / 1\\
Pilot source check & 2 human batches & 95/96 (98.96\%)\\
\addlinespace[2pt]
\multicolumn{3}{@{}l@{}}{\textbf{Third-grader audit \& bounds} ($n = 1{,}445$)}\\
3rd-grader split & P/N/\texttt{U}/miss.\ on 47 P & 37 / 8 / 1 / 1\\
Drop GPT-5.4 & excl.\ shared model & 45/1{,}204 (3.74\%)\\
Count \texttt{U} as P & sensitivity check & 222/1{,}445 (15.36\%)\\
\addlinespace[2pt]
\multicolumn{3}{@{}l@{}}{\textbf{Dual-author blind audit} ($n = 100$)}\\
Flaw $\kappa$ & 3-way / binary & \HAkappaFlaw, \HAkappaBinary\\
Human vs.\ rubric & agree / $\kappa$ & \HAagree\%, \HAkappaHuman\\
Precision / recall & & \HAprec\%, \HArec\%\\
Target miss & wrong-target & \HAtargetMiss\\
Off-key & frame-weighted & \HAoffkey\% \HAoffkeyCI\\
\bottomrule
\end{tabular}

\smallskip
\begin{center}
\footnotesize
\setlength{\tabcolsep}{4.8pt}
\renewcommand{\arraystretch}{1.03}
\begin{tabular}{@{}lrrrr@{}}
\toprule
 & \multicolumn{3}{c}{\textbf{Independent replay label}} & \\
\cmidrule(lr){2-4}
Primary label & Replay P & Replay N & Replay U & Total\\
\midrule
Primary P & $34$ & $0$ & $8$ & $42$\\
Primary N & $0$ & $1{,}062$ & $26$ & $1{,}088$\\
Primary U & $1$ & $118$ & $30$ & $149$\\
\midrule
Total & $35$ & $1{,}180$ & $64$ & $1{,}279$\\
\bottomrule
\end{tabular}

\vspace{1pt}
Replay label matrix ($n = 1{,}279$ reviews with complete pairs).
\end{center}

\end{table}

\noindent\textbf{Reading the visual contrasts.} Table~\ref{tab:visual} and
Figure~\ref{fig:contrasts} bound what the border-highlight interventions can
do. The empirical null is the byte-identical repeat: rerunning the same
model on the same input moves $S_1$ by $+1.39$~pp, $S_2$ by $+0.44$~pp, and
$S_3$ by $-0.22$~pp, all within their bootstrap intervals. Every executed
visual intervention lands inside that same band, except
Target $-$ Random on $S_2$, which shifts by $+5.64$~pp $[+0.83, +10.50]$
with $p = 0.064$ and $p_{\text{Holm}} = 0.767$ after correction across all
twelve contrasts. On the accepted-diagnosis endpoint $S_3$, which is the
endpoint the main paper reports, the interventions move the estimate by at
most $+1.11$~pp with $p = 0.289$. The localization audit confirms that the
edits did what they were meant to do: no target or control edit touched a
caption band, no caption pixel was altered in any of the
$1{,}493{,}559$ altered pixels, and the nearest target edit still sits
$12.5$~pt from its caption. The interventions therefore moved the
measurements rather than the text, and the resulting shifts are of the same
order as the model's own run-to-run noise.

\medskip
\noindent\textbf{Summary of the three checks.} Taken together, the three
blocks above leave the main-text result intact on all three axes. The
rubric-accepted rate is not an artifact of one paper, one reviewer, or one
anchor definition: it stays between $0.83\%$ and $3.25\%$ under every
exclusion and every anchor. It is not an artifact of the rubric alone: a
panel that never saw the cohort reproduces it within $-0.55$~pp, a
source-anchored panel agrees on $79.55\%$ of the 44-item packet, and a
blind human pass confirms roughly half of the sampled acceptances. And it
is not an artifact of how the figure was made salient: every
border-highlight intervention stays inside the empirical repeat band, and
the largest single shift does not survive Holm--Bonferroni correction or
appear at all on the accepted endpoint the paper reports. The one conclusion
the cohort does support is narrow: on the supplied answer key, these models
can be steered to mention the inconsistent element far more often than they
can be steered to assert a diagnosis, and the assertion they do make is
right about half the time.

\begin{table}[!t]
\centering\footnotesize
\setlength{\tabcolsep}{2.9pt}
\renewcommand{\arraystretch}{1.06}
\hyphenpenalty=10000 \exhyphenpenalty=10000
\caption{\textbf{Whole-figure visual highlights yield small shifts within empirical repeat variability and leave rubric-accepted diagnosis statistically indistinguishable from baseline.} Paired visual-intervention contrasts over $K = 31$ papers with $10{,}000$-draw paper-clustered bootstrap 95\% CIs, exact per-paper sign-test $p$, and minimum Holm-Bonferroni $p_{\text{Holm}} = 0.767$ across all 12 contrasts, as detailed in Supplementary Table~\ref{tab:supp_contrasts}.}\label{tab:visual}
\begin{tabular}{@{}lr@{\hspace{8.0pt}}rlr@{}}
\toprule
Contrast & Cells & Diff. & 95\% CI & $p$\\
\midrule
\multicolumn{5}{@{}l@{}}{\textbf{Empirical null: byte-identical repeat}}\\
Orig.\ $-$ Repeat, $S_1$ & 350 & $+1.39$ & {$[-3.65,\ +6.47]$} & 0.678\\
Orig.\ $-$ Repeat, $S_2$ & 350 & $+0.44$ & {$[-4.89,\ +5.57]$} & 0.690\\
Orig.\ $-$ Repeat, $S_3$ & 350 & $-0.22$ & {$[-2.64,\ +2.25]$} & 1.000\\
\addlinespace[2pt]
\multicolumn{5}{@{}l@{}}{\textbf{Executed visual intervention}}\\
Target $-$ Orig., $S_1$ & 349 & $-1.88$ & {$[-6.09,\ +2.57]$} & 0.210\\
Target $-$ Orig., $S_2$ & 349 & $+3.84$ & {$[-0.10,\ +7.67]$} & 0.263\\
Target $-$ Orig., $S_3$ & 349 & $+0.87$ & {$[-0.37,\ +2.49]$} & 0.625\\
Target $-$ Random, $S_1$ & 351 & $+1.19$ & {$[-2.99,\ +5.25]$} & 0.286\\
Target $-$ Random, $S_2$ & 351 & $+5.64$ & {$[+0.83,\ +10.50]$} & 0.064\\
Target $-$ Random, $S_3$ & 351 & $+1.11$ & {$[-0.43,\ +2.71]$} & 0.289\\
Target $-$ Repeat, $S_1$ & 350 & $-0.92$ & {$[-5.54,\ +3.71]$} & 0.845\\
Target $-$ Repeat, $S_2$ & 350 & $+4.19$ & {$[+0.10,\ +8.20]$} & 0.134\\
Target $-$ Repeat, $S_3$ & 350 & $+0.65$ & {$[-2.15,\ +3.49]$} & 1.000\\
\addlinespace[2pt]
\multicolumn{5}{@{}l@{}}{\textbf{Localization audit of the 62 executed edits}}\\
\multicolumn{5}{@{}l@{}}{Target / control edits touching a caption band: 0/31 and 0/30}\\
\multicolumn{5}{@{}l@{}}{Altered pixels inside any caption band: 0 / $1{,}493{,}559$}\\
\multicolumn{5}{@{}l@{}}{Nearest target edit to its caption: 12.5~pt, median 49.5~pt}\\
\bottomrule
\end{tabular}

\end{table}

\medskip
\noindent\textbf{What these checks do not establish.} Three limits are worth
stating, because the appendix is easy to read as a stronger claim than it
is. First, every check here is \emph{verification}, not \emph{discovery}: the
cohort ships a registered answer key for all 31 papers, so the models are
scored against a label that already exists, and nothing here shows that
they can find an unannotated inconsistency. Second, the agreement figures
are agreement \emph{with the registered key}, and the human audit shows that
roughly half of the accepted reviews do not carry the same diagnosis that
the rubric assigned; where the audit and the rubric disagree, the audit is
the better reader of the manuscript, so the $3.25\%$ headline counts some
diagnoses that a careful human would not make. The bias is therefore toward
over-counting, and it is why the main paper reports the rate as an upper
bound on verified diagnosis rather than a detection rate. Third, the checks
bound the instrument but not the construct: they say nothing about whether
the benchmark's notion of inconsistency matches the one a domain expert
would use, and the human audit in Table~\ref{tab:graders} is the only probe
here that touches it.

\end{document}